\documentclass[lettersize,journal]{IEEEtran}
\usepackage{url}
\usepackage{xcolor}
\usepackage[shortlabels]{enumitem}
\usepackage{graphicx} % 图片功能包
\usepackage{subfigure}	% 子图功能包
\usepackage{cite}
\usepackage{hyperref}
\usepackage{multirow}
\usepackage{caption}
\hypersetup{hypertex=true,
	colorlinks=true,
	linkcolor=blue,
	anchorcolor=blue,
	citecolor=blue,
	urlcolor=blue}
\usepackage{amsmath}
\usepackage[ruled]{algorithm2e}
\usepackage{amssymb}
\usepackage{booktabs}
\usepackage{soul}
\usepackage{colortbl}
\soulregister\cite7 % 针对\cite命令
\soulregister\ref7 % 针对\ref命令
\usepackage{adjustbox}
\usepackage{cuted}
\usepackage{lineno}

\definecolor{myrevise}{RGB}{255,0, 0}
\begin{document}

\title{3D-UIR: 3D Gaussian for Underwater 3D Scene Reconstruction via Physics-Based Appearance-Medium Decoupling}

\author{Jieyu Yuan, 
			Yujun Li, 
			Yuanlin Zhang, 
			Chunle Guo,~\IEEEmembership{Member,~IEEE}, \linebreak
			Xiongxin Tang, 
			Ruixing Wang, 
			and Chongyi Li,~\IEEEmembership{Senior Member,~IEEE}
        % <-this % stops a space
\thanks{This work was supported in part by the National Natural Science Foundation of China (62501312), the China Postdoctoral Science Foundation (2025M781478), the Major Science and Technology Project on Artificial Intelligence of Tianjin (No. 25ZXRGGX00290), the China Postdoctoral Science Foundation - Tianjin Joint Support Program (2025T022TJ), and the Tianjin Natural Science Foundation Project (25ZXRGGX00290, 24JCJQJC00020, 25JCQNJC01390), and the Fundamental Research Funds for the Central Universities (Nankai University, 63243143, 63253223, 63253219). (\emph{Corresponding author: Chongyi Li.})}
\thanks{Jieyu Yuan, Yujun Li and Yuanlin Zhang are with the VCIP, College of Computer Science, Nankai University, Tianjin 300350, China (e-mail: jieyuyuan.cn@gmail.com; yujunli@mail.nankai.edu.cn; zason\_zyl@163.com).}
\thanks{Xiongxin Tang is with the Institute of Software, Chinese Academy of Sciences, Beijing 100190, China (e-mail: xiongxin@iscas.ac.cn).}
\thanks{Ruixing Wang is with DJI, Shenzhen 518000, China (e-mail: ruixingw@hustunique.com).}
\thanks{Chunle Guo and Chongyi Li are with the VCIP, College of Computer Science, Nankai University, Tianjin 300350, China, and also with the Nankai International Advanced Research Institute (NKIARI), Shenzhen Futian, China (e-mail: guochunle@nankai.edu.cn; lichongyi@nankai.edu.cn).}
}

\markboth{Journal of \LaTeX\ Class Files,~Vol.~14, No.~8, August~2021}%
{Shell \MakeLowercase{\textit{et al.}}: A Sample Article Using IEEEtran.cls for IEEE Journals}


\maketitle

\begin{strip}
    \centering
    \vspace{-4.5cm}
    \includegraphics[width=\textwidth]{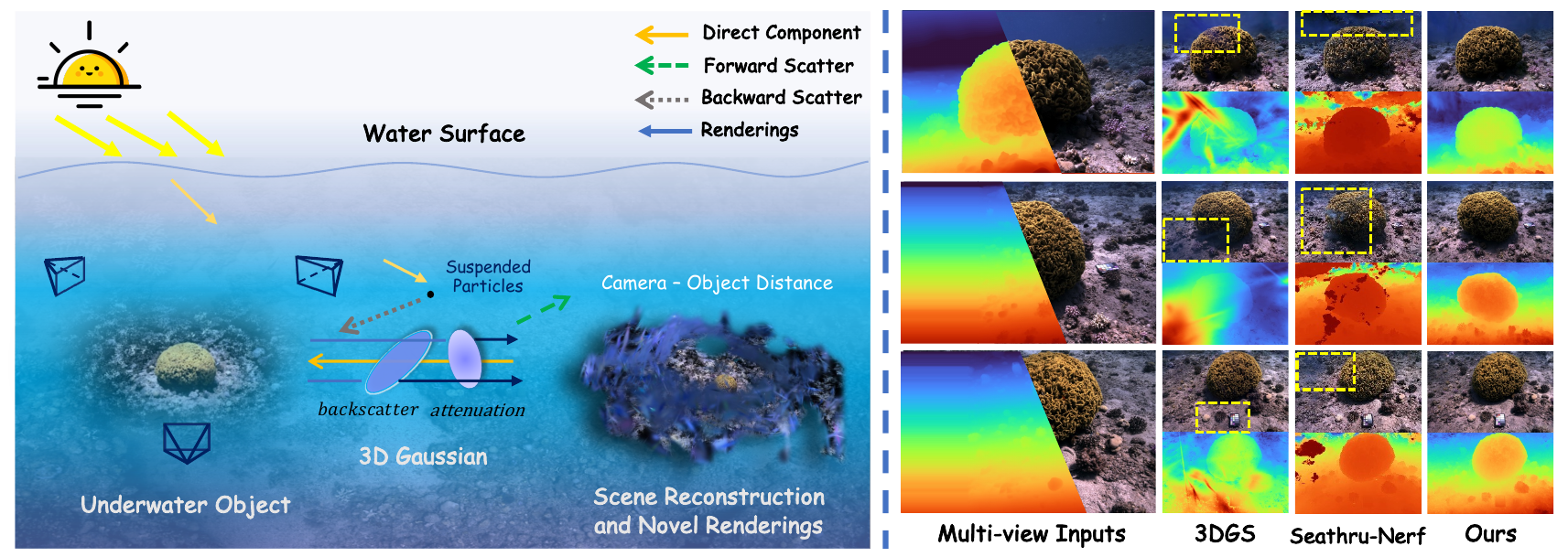}
    \captionof{figure}{Underwater optical scattering and absorption pose unique challenges for novel view synthesis: the standard volume rendering equation inadequately models participating media, resulting in floating artifacts (\textbf{left}), while wavelength-dependent attenuation causes inconsistent appearance across viewpoints (\textbf{right}). 3DGS \cite{3DGS_2023} lacks scattering media modeling, leading to water column effects and collapsed depth. SeaThru-NeRF \cite{SeaThru-NeRF_2023_CVPR} fails to handle dynamic photometric variations, introducing water float artifacts (\colorbox[rgb]{1.0, 1.0, 0.8}{yellow boxes}). Our approach explicitly models the participating medium, achieving photorealistic novel views with consistent rendering and effective artifact elimination.}
    \vspace{-0.2cm}
    \label{fig:teaser}
\end{strip}

\begin{abstract}
Novel view synthesis for underwater scene reconstruction presents unique challenges due to complex light-media interactions. Optical scattering and absorption in water body bring inhomogeneous medium attenuation interference that disrupts conventional volume rendering assumptions of uniform propagation medium. While 3D Gaussian Splatting (3DGS) offers real-time rendering capabilities, it struggles with underwater inhomogeneous environments where scattering media introduces artifacts and inconsistent appearance. In this study, we propose a physics-based framework that disentangles object appearance from water medium effects through tailored Gaussian modeling. Our approach introduces appearance embeddings, which are explicit medium representations for backscatter and attenuation, enhancing scene consistency. In addition, we propose a depth-guided optimization strategy that leverages pseudo-depth maps as supervision with depth regularization and scale penalty terms to improve geometric fidelity. By integrating the proposed appearance and medium modeling components via an underwater imaging model, our approach achieves both high-quality novel view synthesis and physically accurate scene restoration. Experiments demonstrate our significant improvements in rendering quality and restoration accuracy over existing methods. The project page is available at \href{https://bilityniu.github.io/3D-UIR}{https://bilityniu.github.io/3D-UIR}.
\end{abstract}

%-------------------------------------------------------------------------


%-------------------------------------------------------------------------
\section{Introduction}

\IEEEPARstart{U}{nderwater} optical images are crucial for marine information capture, as their rich textures and accessibility make them indispensable for a wide range of underwater vision applications \cite{Zhang_2024_CVPR,yuan_tii,Zhou2025}.
Despite these advantages, planar 2D imaging inherently suffers from limited spatial information representation, which is insufficient for the precision modeling and advanced analytical requirements of modern marine applications.
The transition from 2D imagery to volumetric 3D models enables substantially richer geometric characterization and spatial context in underwater 3D reconstruction \cite{Matan_2024}, making it a key technology for marine mapping, underwater archaeology, environmental monitoring, and autonomous underwater vehicle navigation.
Multi-view imagery enables precise spatial analysis of underwater environments \cite{Wang_ICRA_2023}, and supports scene comprehension in stereoscopic 3D space to support more complex task requirements.

Unlike clear-air environments, underwater scenes present unique optical challenges due to complex light-medium interactions. 
Specifically, light propagation through water involves wavelength-dependent attenuation and particle scattering \cite{uieb,Akkaynak_2017_CVPR,Li_2016_tip}, resulting in depth-dependent color distortion and visibility degradation \cite{Zhang_tgrs}.
These optical phenomena result in both forward and backward scattering effects, creating veiling light and spatial blur that significantly degrade underwater image quality.
The physical constraints of underwater environments alter image formation, introducing greenish or bluish color casts and blurred backgrounds that pose challenges for general scene reconstruction methods.
Consequently, underwater degradation fundamentally disrupts the photometric consistency assumptions essential for multi-view 3D reconstruction.

The unique optical properties of underwater environments challenge existing volume rendering-based view synthesis methods, which typically assume light transport involves only emission and absorption while neglecting underwater scattering and attenuation effects. While recent advances \cite{nerf_2020_ECCV, 3DGS_2023} have demonstrated remarkable success in general scenes, they face fundamental limitations underwater. Specifically, alpha blending models radiance propagation through emission and absorption only \cite{nerf_2020_ECCV, DehazeNeRF_2024, 3DGS_2023}, while underwater scenarios involve continuous in-scattering and attenuation from the participating water medium \cite{Akkaynak2017}, introducing translucent effects absent in clear-air rendering.
This creates \textbf{two key challenges}: \textbf{1) Medium Veiling Opacity:} continuous scattering along light paths results in depth-dependent veiling effects that cannot be accurately represented through standard opacity models \cite{Sea-Thru_2019_CVPR}. \textbf{2) View-Dependent Inconsistency:} depth-dependent attenuation causes significant appearance variations across viewpoints, violating the consistent radiance assumption \cite{NeRF_in_Wild_2021_CVPR}. In underwater environments, the same scene point exhibits varying colors from different viewpoints due to wavelength-dependent attenuation along different water path lengths. 
Recent underwater NVS methods \cite{SeaThru-NeRF_2023_CVPR, watersplatting_2024, SeaSplat_2024} incorporate scattering media modeling but often rely on general-purpose appearance representations that lack explicit handling of these view-dependent variations. 
As illustrated in Fig. \ref{fig:teaser}, this entanglement leads to water-cloud floats, view-dependent artifacts, and photometric inconsistencies in novel views.

In this study, we propose 3D-UIR (3D Gaussian for Underwater Image Reconstruction and Restoration), a physics-based underwater scene reconstruction framework that explicitly models the complex light transport characteristics in underwater environments.
Our approach employs a 3D Gaussian representation to decompose underwater scenes by separating intrinsic scene properties from water-induced effects.
Specifically, we introduce a dual-branch architecture: an object appearance branch that leverages pose-based embeddings to model camera-specific radiance variations, and a medium appearance branch that explicitly parameterizes scattering and attenuation. Unlike methods relying on per-image indices or implicit neural networks, our pose-based design enables direct generalization to novel viewpoints by continuously modeling the relationship between viewing geometry and attenuation. We further incorporate depth-guided optimization with pseudo-depth supervision to jointly refine geometry and medium parameters. This comprehensive approach enables both high-fidelity novel view synthesis and scene restoration by parametrically removing water medium effects.

In summary, we make the following contributions:
\begin{itemize}  
	\item 	\textbf{Decoupled Underwater Representation:} We propose a physics-based underwater 3DGS representation that decouples object appearance from water medium effects, explicitly modeling complex light transport in participating media while removing water float artifacts caused by translucent scattering media.
    \item \textbf{View-Consistent Scene Restoration:} We propose an underwater appearance model that integrates geometry-aware features with pose-based embeddings to capture depth-dependent color variations, enabling consistent scene restoration across novel viewpoints without reference image supervision. A pixel-based white balance technique further ensures optimal color balance while preventing oversaturation.
	\item   \textbf{Depth Guided Optimization:} We propose a distance optimization strategy that establishes full gradient coupling between geometry and medium estimation by using rendered depth as direct input to scatter modeling, enabling joint optimization that reduces volumetric artifacts while achieving physically accurate reconstruction.
\end{itemize}

%-------------------------------------------------------------------------
\section{Related Work}
%-------------------------------------------------------------------------
\subsection{Novel View Synthesis}

Novel View Synthesis generates arbitrary views of a scene from limited 2D images. Neural Radiance Field (NeRF) \cite{nerf_2020_ECCV} maps color and density to an implicit function, enabling high-quality rendering through volume rendering \cite{nerf_2020_ECCV}. 
In contrast, 3D Gaussian Splatting (3DGS) \cite{3DGS_2023} uses explicit representations, achieving superior real-time performance by representing scene points as Gaussian primitives with a differentiable tile-based rasterizer \cite{3DGS_2023,Mip-Splatting_2024_CVPR,absgs_acmmm_2024}.
While both methods achieve high-fidelity reconstructions \cite{Mildenhall_2022_CVPR, hdr_gs, Deblurring-GS, LE3D_2024}, they struggle in scenarios with inconsistent scene density and radiance, often leading to ghosting and artifacts in unconstrained image collections \cite{NeRF_in_Wild_2021_CVPR, SWAG_eccv_2024}.
To address this issue, recent works model inconsistent components separately.
For example, NeRF-W \cite{NeRF_in_Wild_2021_CVPR} introduces per-image latent appearance embeddings to decompose scenes into image-dependent and shared components. 
SWAG \cite{SWAG_eccv_2024} extends appearance modeling by modulating Gaussian colors and constructing image-dependent opacity variations. 
Similarly, GS-W \cite{GS-W_eccv_2024} leverages distinct intrinsic and dynamic appearance features for complex scenes.
Underwater scenes share similarities with ``in-the-wild" scenes in terms of medium-object interactions. 
Inspired by appearance modeling techniques for variable scene conditions, we propose a novel approach that incorporates latent appearance modeling to separately represent object and scattering medium characteristics, integrating an underwater physical imaging model to improve reconstruction consistency.

\subsection{Underwater Scene Representation}

Underwater novel view synthesis presents unique challenges due to the participating media. While standard methods assume clear air, underwater approaches must account for wavelength-dependent attenuation and backscatter.
SeaThru-NeRF \cite{SeaThru-NeRF_2023_CVPR} pioneered this direction by integrating the comprehensive underwater image formation model (UIFM) into the NeRF rendering pipeline, separating direct and backscatter components. 
Similarly, WaterHE-NeRF \cite{WaterHE-NeRF_2024_IF} proposes a water ray-tracing field that encodes color, density, and illumination attenuation in 3D space based on Retinex theory.
However, its reliance on implicit neural networks leads to slow training and inference.

To achieve real-time performance, recent works have adopted 3DGS as the new representation. Water-Splatting \cite{watersplatting_2024} and SeaSplat \cite{SeaSplat_2024} extend 3DGS with depth-aware medium modeling, but they fundamentally rely on per-image appearance embeddings \cite{NeRF_in_Wild_2021_CVPR} to handle color variations. While effective for training views, these embeddings fail to generalize to novel viewpoints without reference images. Similarly, UW-GS \cite{UW-GS_WACV_2025} and Aquatic-GS \cite{Aquatic-GS} employ implicit MLPs to estimate medium parameters or construct neural water fields.  More recent explorations include DualPhys-GS \cite{DualPhysGS}, which proposes dual-path optimization for attenuation and scattering with scene-adaptive mechanisms, and UW-3DGS \cite{UW_3DGS}, which introduces physics-aware uncertainty pruning to reduce floating artifacts.
While these methods improve geometric accuracy via depth guidance, their reliance on implicit networks increases overhead and limits the ability to capture global, depth-dependent color shifts in underwater imaging.

Our method advances this line of research by proposing a fully explicit and decoupled framework. Unlike prior works that use implicit networks or per-image indices, we introduce pose-based appearance embeddings and explicit medium parameterization. This design allows our model to capture view-dependent attenuation physically and generalize naturally to unseen viewpoints, effectively bridging the gap between real-time rendering and physical accuracy.

\subsection{Underwater Image Restoration}
Underwater image restoration aims to recover clear visual perception from degraded imagery \cite{Li_survey_2020, Shuang2024}. Existing approaches typically fall into two categories: image enhancement methods and physics-based restoration. Enhancement-based methods \cite{ucolor,yuan2025w2wdiff} directly process image pixels to improve perceptual quality, such as contrast, saturation, and sharpness, often without explicit physical modeling. Conversely, physics-based methods \cite{Liang2022, PUGAN, Osmosis_eccv_2024} invert the Underwater Image Formation Model (UIFM) \cite{Akkaynak_2018_CVPR} to estimate attenuation coefficients and recover the undistorted scene radiance. While these methods can produce realistic results, single-image restoration is inherently ill-posed and typically requires strong prior assumptions \cite{Sea-Thru_2019_CVPR} to constrain the solution space. 
In the context of 3D reconstruction, multi-view consistency serves as a powerful constraint to resolve the ambiguities inherent in single-image restoration. By leveraging geometric alignment and redundant observations across multiple viewpoints, our framework enables robust estimation of physical parameters without relying on strong hand-crafted priors. This allows our method to achieve physically accurate scene restoration as an intrinsic outcome of the reconstruction process, ensuring global consistency across all viewing angles.

%-------------------------------------------------------------------------
\section{Preliminaries}

\subsection{3D Gaussian Splatting}
3DGS \cite{3DGS_2023} is an explicit representation technique that offers efficient, high-quality rendering with real-time performance. The method initializes 3D Gaussians from Structure-from-Motion (SfM) \cite{SFM_2016_CVPR} points, transforming them into volumetric primitives that are projected onto 2D space. Each 3D Gaussian is defined as:
\begin{equation}
	G(x - \mu, \Sigma) = \exp({-\frac{1}{2} (x - \mu)^T \Sigma^{-1} (x - \mu)}),
\end{equation}
where $\mu \in \mathbb{R}^3$  is the center position of the Gaussian points, $\Sigma$ is a 3D covariance matrix that represents the shape, scale, and orientation characteristics of the Gaussian primitives in 3D space.
To ensure $\Sigma$ remains positive semi-definite during optimization, it is factorized into a rotation quaternion $ \in \mathbb{R}^4$ and a scaling factor $\in \mathbb{R}^3$:
\begin{equation}
	\Sigma = R S S^T R^T.
\end{equation}

3DGS employs a tile-based rasterizer to efficiently render the scene by projecting all Gaussians onto the image plane. Each Gaussian carries additional attributes, including opacity ($\alpha$) and color ($c$), with color represented using third-order spherical harmonic coefficients to model view-dependent appearance.
The final color for each pixel is computed through $\alpha$-blending by accumulating the contributions from overlapping 2D Gaussians:
\begin{align}
C = \sum_{i \in G_r} c_i \sigma_i \prod_{j=1}^{i-1} (1 - \sigma_j), \quad
\sigma_i = G_r(px' - \mu_i, \Sigma'_i),
\end{align}
where $r$ represents the position of a pixel and $G_r$ denotes the sorted Gaussian points associated with that pixel.
Similarly, depth maps are rendered by accumulating the depth values $d_i$ of ordered Gaussian primitives along each ray:
\begin{equation}
	\mathcal{D}=\sum_{i\in n_i}d_i\alpha_i\prod_{j=1}^{i-1}(1-\alpha_j).
    \label{eq: depth}
\end{equation}

All Gaussian attributes are optimized via the differentiable rasterizer, with 3DGS dynamically splitting and duplicating Gaussians based on size and gradient magnitude to enhance reconstruction fidelity.

\begin{figure*}[!t]
	\centering
	\includegraphics[width=0.96\linewidth]{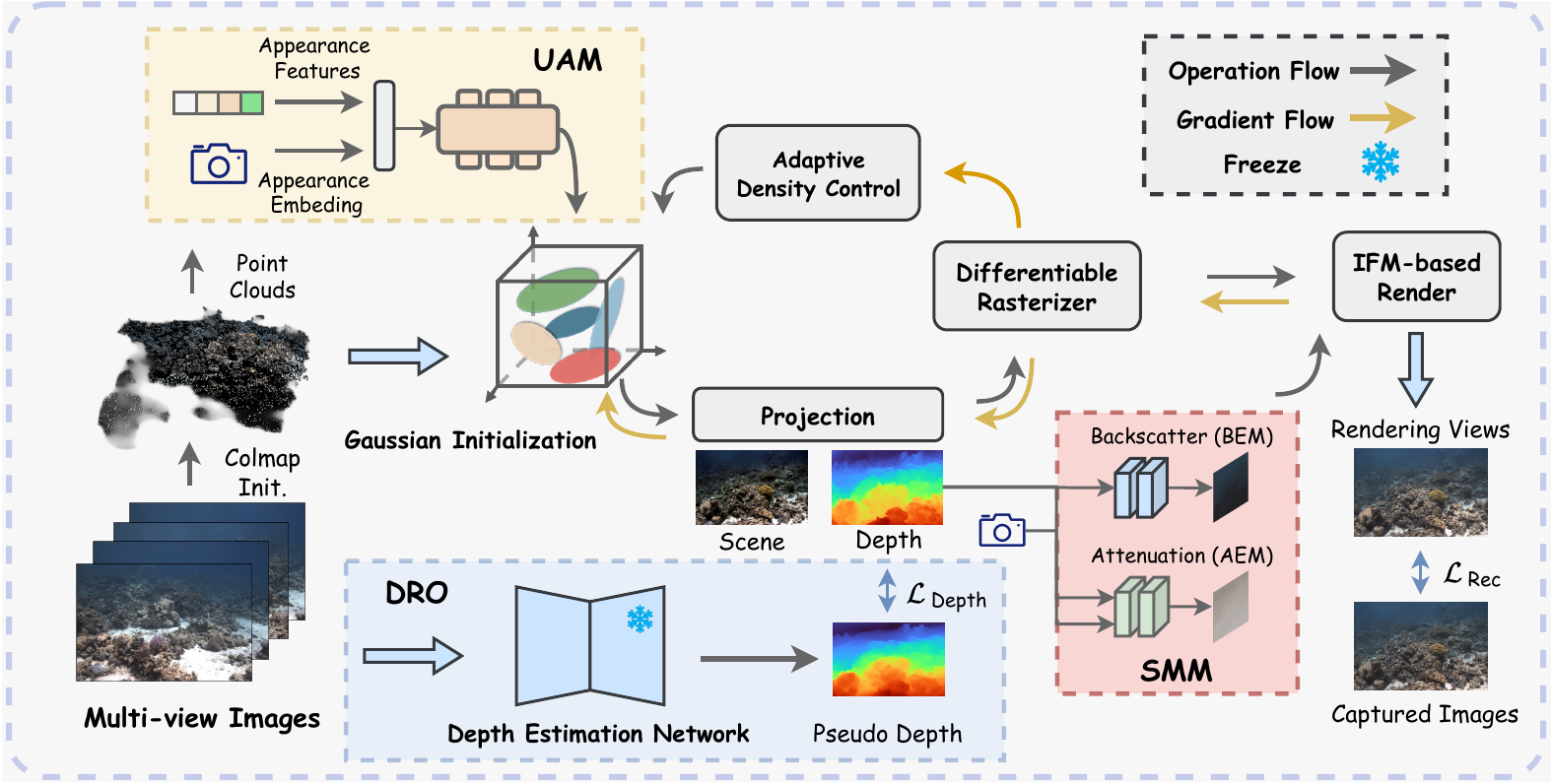}
	\caption{
    Overview of the proposed method. Our method disentangles object appearance from water medium effects. The \textcolor{yellow!70!black}{Underwater Appearance Modeling} (\textcolor{yellow!70!black}{\textbf{UAM}}) branch models view-consistent appearance via pose-based embeddings. The \textcolor{red}{Scatter Medium Modeling} (\textcolor{red}{\textbf{SMM}}) branch estimates backscatter and attenuation from rendered depth. The \textcolor{blue!70}{Depth-guided Regularization Optimization} (\textcolor{blue!60}{\textbf{DRO}}) leverages pseudo-depth maps for geometric supervision. All components are jointly optimized via differentiable rasterization through a physics-based underwater image formation model.}
	\label{fig:pipeline}
\end{figure*}

\subsection{Underwater Image Formation Model}
Underwater imagery suffers from degradation due to scattering and absorption during light propagation, introducing attenuation and backscatter effects that impair information representation \cite{Akkaynak_2017_CVPR}. The captured image $ I$ intensity combines two components: direct object radiance traveling to the observer and backscatter from all directions along the light path.
Following the revised underwater image formation model \cite{Akkaynak_2018_CVPR,Sea-Thru_2019_CVPR}, the scene captured along line-of-sight distance $z$ is expressed as:
\begin{equation}
    I=J \cdot e^{-\beta^D \cdot z}  + B^{\infty} \cdot \left( 1 - e^{-\beta^B \cdot z} \right), 
    \label{eq:UIFM}
\end{equation}
where  $J$ is the unattenuated image from the ideal clear scene and $B^{\infty}$ is the value of backscatter at infinite distance. Here, $\beta^D$ and $\beta^B$ are the attenuation coefficients for direct and backscatter components, respectively. 
Their coefficients depend on the optical properties of the medium, such as water turbidity, suspended particles, and wavelength, and thus can vary significantly across different underwater environments.
UIFM assumes objects are opaque while the medium is semi-transparent with constant low density~\cite{Akkaynak_2018_CVPR}. This distinction enables separate modeling of backscatter effects and object radiance. However, in neural representations, direct light reflection and translucent media often appear ambiguous when modeled as Gaussian primitives \cite{MiniSplatting_eccv_2024,DehazeNeRF_2024}. To address this challenge, our method separately models objects and media, computing attenuation parameters based on UIFM to enhance underwater scene reconstruction quality.

\section{Methodology}

Underwater scattering medium poses significant challenges to conventional 3D GS, introducing volumetric artifacts and compromising scene consistency due to complex light interactions and depth-dependent attenuation that varies across viewpoints. We propose a physics-based method to address these challenges. 

Our approach begins with multi-view underwater images processed through COLMAP \cite{SFM_2016_CVPR} for Gaussian primitive initialization. After creating 3D Gaussians, Underwater Appearance Modeling (UAM) encodes appearance features and view embeddings derived from camera poses through an MLP that applies affine color transformations. These per-Gaussian colors undergo 2D projection and tile-based rasterization to generate an initial scene appearance representation.
At the same time, Scatter Medium Modeling (SMM) estimates view-specific backscatter and attenuation parameters based on the projected scene depth information. 
The integration of these branches occurs in the IFM-based rendering stage, where clean scene colors from the projection step serve as input to our Underwater Image Formation Model (UIFM). The UIFM combines these colors with the medium parameters estimated by the SMM branch through physics-based $\alpha$-blending, transforming clean rendered colors into degraded underwater images that match the photometric characteristics of captured underwater images for training supervision.
Additionally, the Depth-guided Regularization Optimization (DRO) provides pseudo-depth supervision to improve parameter estimation accuracy. During training, Adaptive Density Control dynamically manages the Gaussian distribution through cloning, pruning, and strategic addition of primitives, which helps balance scene fidelity and computational efficiency.
The complete framework is illustrated in Fig. \ref{fig:pipeline}.

\subsection{Underwater Appearance Modeling}

Underwater imagery suffers from distance-dependent attenuation, causing inconsistent colors and photometric variations across viewpoints, which severely compromise both novel view synthesis and scene restoration quality.
To address these issues, we introduce an underwater appearance modeling framework.
By leveraging specialized per-Gaussian appearance embeddings, our method effectively captures local variations and mitigates view-dependent color distortions inherent to the underwater medium.
We adopt an adaptive affine color transformation that explicitly models view-specific conditions.
Specifically, for the $i$-th Gaussian primitive observed from viewing direction $\mathbf{r}$, we parameterize the transformation using channel-wise scaling and offset factors $(\beta,\gamma)=\{(\beta_{k},\gamma_{k})\}_{k=1}^{3}$ for each color channel $k$:
\begin{equation}
	\tilde{\mathbf{c}}_i = \gamma \cdot \hat{\mathbf{c}}_i(\mathbf{r}) + \beta,
\end{equation}
where $\hat{\mathbf{c}}_i(\mathbf{r})$ represents the view-dependent base color and $\tilde{\mathbf{c}}_i$ denotes the transformed result.

For in-the-wild scenes, NVS methods \cite{NeRF_in_Wild_2021_CVPR, GS-W_eccv_2024, SWAG_eccv_2024} employ trainable per-image index embeddings to represent appearance variations. During inference, these methods require ground truth images, often using one half for appearance matching while evaluating metrics on the other half. However, this reliance on per-input view supervision poses a fundamental limitation for underwater environments, where training viewpoints are significantly sparser than the novel view requirements.
To overcome this limitation, we construct appearance embeddings from camera poses, exploiting the insight that camera parameters optimized through global bundle adjustment adhere to a consistent maximum a posteriori probability distribution \cite{Scaffold-SLAM}. 
This formulation enables effective generalization to novel viewpoints without requiring training on the test set.
We implement this approach through a tiny MLP (two-layer MLP with 128 hidden units) that projects camera parameters into a low-dimensional appearance latent space to capture the underlying probabilistic distribution of these poses:
\begin{equation}
	\mathbf{e}_i=\mathrm{MLP}(\varphi\{\mathbf{R}, \mathbf{t}\}),
\end{equation}
where $\varphi\{\cdot\}$ denotes positional encoding, and $\mathbf{R}$ and $\mathbf{t}$ correspond to the rotation matrix and translation vector of the camera pose, respectively.

To further enhance appearance consistency while preserving spatial coherence, we associate each Gaussian with a trainable appearance feature $\mathbf{f}_i$. Rather than employing random initialization that would lack intrinsic spatial correlation, we initialize these features using multi-scale Fourier encoding to establish a locality-preserving prior:
\begin{equation}
	\mathbf{f}_i = \text{Concat}\{[\sin(\pi p_k 2^m), \cos(\pi p_k 2^m)]\},
\end{equation}
where point coordinates $p_k$ (for $k \in \{1,2,3\}$) are normalized to $[0,1]$ using the $0.97$ quantile of the $L^{\infty}$ norm, with $m$ ranging from $1$ to $4$, following the setting in \cite{2024wildgaussians}.

Our complete appearance modeling framework uses a parameterized function $\mathcal{F}_\theta$ to synthesize the image-conditioned color by combining the base Gaussian color $\mathbf{c}_i$, appearance feature $\mathbf{f}_i$, and view-specific embedding $\mathbf{e}_j$:
\begin{equation}
	\hat{\mathbf{c}}_{i} = \mathcal{F}_\theta(\mathbf{c}_i, \mathbf{f}_i, \mathbf{e}_j).
\end{equation}

This formulation integrates both geometric information through position-encoded features $\mathbf{f}_i$ and view-dependent variations via pose-based embeddings $\mathbf{e}_j$.
Unlike per-image index embeddings \cite{NeRF_in_Wild_2021_CVPR,2024wildgaussians,Splatfacto-W_2024} that require reference image supervision, our pose-based approach enables direct generalization to novel viewpoints by modeling distance-dependent attenuation through the continuous camera pose space. This is beneficial for sparse underwater captures, where our model maintains a consistent appearance across diverse viewpoints while effectively disentangling intrinsic scene properties from water-induced color variations.

\begin{figure*}[th]
	\centering
	\includegraphics[width=0.96\linewidth]{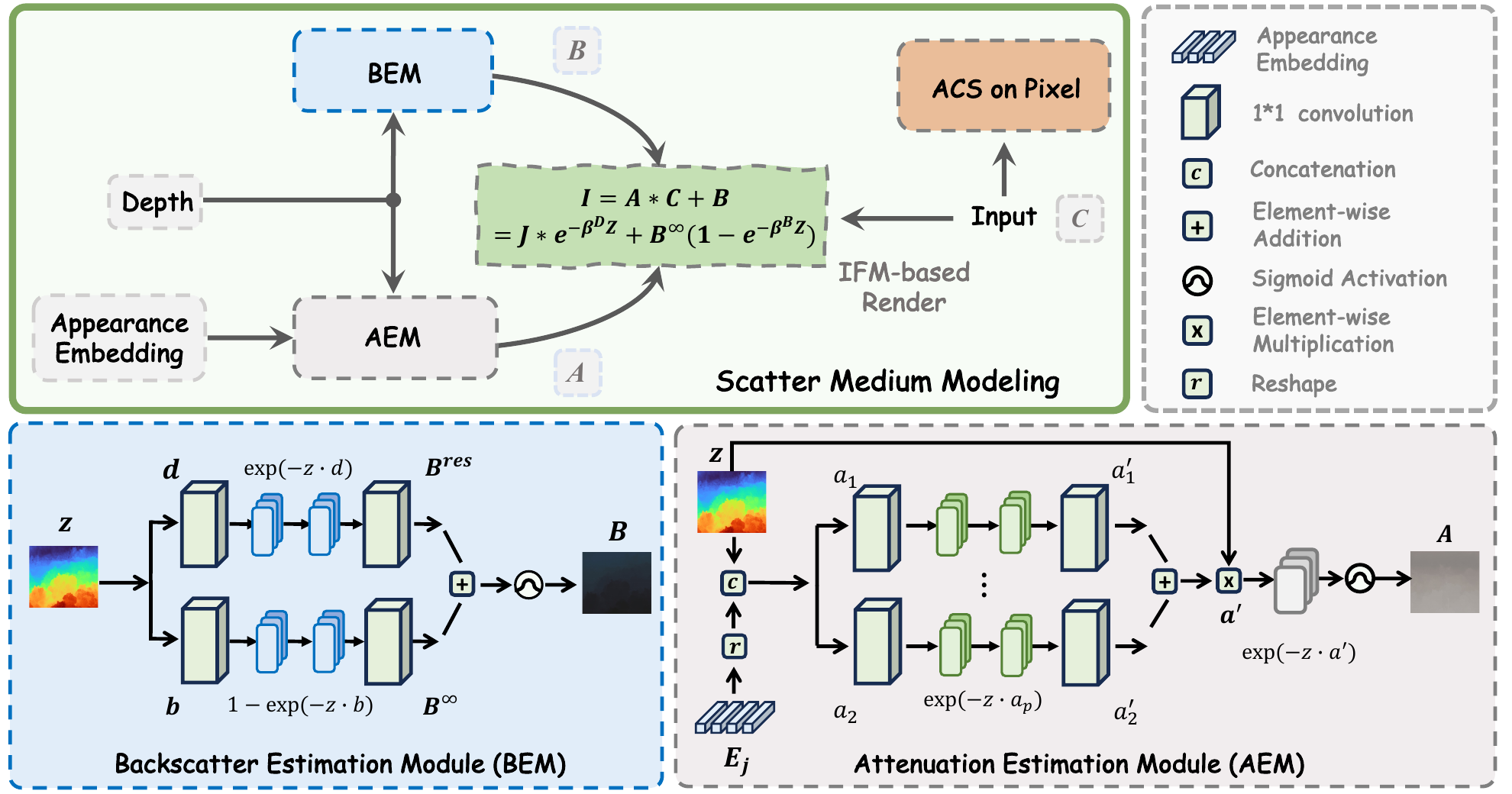} % 适应两栏宽度
	\caption{The top illustrates an overview of the proposed \textbf{Scatter Medium Modeling} framework, which decomposes underwater medium effects based on a physical imaging model. The bottom presents the detailed architectures of the \textbf{Backscatter Estimation Module (BEM)} and the \textbf{Attenuation Estimation Module (AEM)}, respectively. The BEM computes the backscatter component, while the AEM estimates the attenuation coefficients of the direct component, effectively modeling the complex light transport in underwater environments.} 
	\label{scatter_model}
\end{figure*}

\subsection{Scatter Medium Modeling}

Underwater scenes exhibit floating artifacts from backscatter, which intensifies with distance independently of scene content. Recent works \cite{SeaThru-NeRF_2023_CVPR,watersplatting_2024,UW-GS_WACV_2025} employ Neural Water Fields to implicitly estimate medium parameters through point-based features and radiance MLPs. However, this implicit approach suffers from initialization sensitivity and increased computational complexity due to viewpoint-conditioned MLPs, while neglecting the fundamental distance-dependent distribution of underwater light transport parameters.

To address these limitations, we propose a depth-guided explicit scene medium modeling approach that decomposes underwater medium effects into backscatter and attenuation estimations. 
Unlike the implicit mappings used in recent neural methods or the linear fitting procedures adopted by Sea-thru~\cite{Sea-Thru_2019_CVPR}, our framework introduces a lightweight, learning-based approach that explicitly models scattering media. Built upon the physical principles of Sea-thru, the architecture of our \textbf{Scatter Medium Modeling} framework is illustrated in Fig.~\ref{scatter_model}.
This physically principled approach enhances the representation capacity for complex underwater light transport phenomena through a fully convolutional structure that directly processes depth information to optimize medium parameters \cite{DeepSeeColor}. In comparison to the traditional MLP-based or linear fitting optimization approaches, our method significantly reduces computational complexity while maintaining high efficiency.

For backscatter estimation, we formulate the backscatter intensity $\hat{B}_c$ based on
\cite{Sea-Thru_2019_CVPR} for color channel $c$ as:
\begin{equation}
	\hat{B}_{c} = B_{c}^{\infty}\left(1-\exp\left(-b\cdot z\right)\right)+B_{c}^{res}\exp\left(-d\cdot z\right),
\end{equation}
where $B_{c}^{\infty}$ denotes the asymptotic backscatter intensity at infinite distance, $b$ represents the backscatter accumulation rate, $z$ corresponds to the distance in water, and $B_{c}^{res}$ and $d$ are modulation parameters that characterize near-distance backscatter behavior. To efficiently implement this physics-based model, we employ convolutional operations to process the rendered depth map $D$, enabling direct estimation of the relevant optical parameters. The parameters are computed through 2D convolutional layers to generate the coefficients $b$ and $d$ for each color channel, which allows these parameters to be optimized through gradient descent during training.

For attenuation estimation, we utilize an exponential function to approximate the attenuation coefficient $a_c(z)$ for each color channel \cite{Sea-Thru_2019_CVPR}:
\begin{equation}
	a_c(z)=\sum_{p=1}^P a_c^{p\prime}\exp(-z \cdot a_c^p),
\end{equation}
where $a_c^{p\prime}$, $a_c^p \in [0, 1]$, and $P$ represent the number of exponential functions used to approximate the complex wavelength-dependent attenuation, respectively.

To mitigate dependency on initial parameter values and enhance robustness, we incorporate depth information and appearance embedding $\mathbf{e}$ as initialization context for medium parameter estimation. Specifically, we concatenate the depth map $D$ with the appearance embedding $\mathbf{e}_j$ from our UAM branch to construct a comprehensive representation of scene characteristics:
\begin{equation}
	z' = \text{Concat}(D, \mathbf{e}_j).
\end{equation}

After estimating the medium parameters, our comprehensive underwater image formation model integrates both appearance color $\hat{C}$ and medium properties through a physics-based formulation. Through the joint optimization of this model, we can simultaneously obtain the reconstructed underwater image and recover the true color without water effects following Eq~\eqref{eq:UIFM}:
\begin{equation}
	I_c = \underbrace{a_c(z') \cdot \hat{C}_c}_{\text{Direct}} + \underbrace{\hat{B}_{c}}_{\text{Backscatter}}.
	\label{eq:IFM}
\end{equation}

The rendered restored image $C$ represents colors corrected along the depth direction. Depending on the imaging geometry, this intermediate result may require additional correction to achieve color characteristics equivalent to those of an image captured at the sea surface \cite{Sea-Thru_2019_CVPR}.
General underwater restoration approaches \cite{Sea-Thru_2019_CVPR,PUGAN} employ global or local white balancing post-processing on the restored image $J$ to normalize color distribution and improve visual appearance.
Similar to these underwater image restoration works, we apply adaptive contrast stretching (ACS) on the rendered image $C$ to avoid color oversaturation. The mapping process is formulated as:
\begin{equation}\label{mapping}
I^c_{tm} = (I^c - I^c_{\mathrm{min}})\frac{I^o_{max}-I^o_{min}}{I^c_\mathrm{max} - I^c_{\mathrm{min}}},
\end{equation}
where $I^o_{max}$ and $I^o_{min}$ define the target stretching ranges, as measured by the Gray-world hypothesis \cite{CBM_2021_tgrs}, ensuring optimal color balance and contrast enhancement.

By leveraging rendered depth maps for medium parameter optimization and incorporating physical constraints into the 3D Gaussian representation, we effectively disentangle water effects from object appearance, significantly reducing floating artifacts while enhancing cross-view consistency and physical accuracy in underwater 3D reconstruction.

\subsection{Reconstruction Optimization}

Our underwater 3D GS framework employs reconstruction loss ($\mathcal{L}_r$), depth regularization loss ($\mathcal{L}_d$), and underwater scale regularization loss ($\mathcal{L}_s$) to reconstruct complex underwater scenes. The overall loss function $\mathcal{L}$ is defined as:
\begin{equation}
	\mathcal{L}= \mathcal{L}_r +  \mathcal{L}_d + \mathcal{L}_s.
\end{equation}

For the reconstruction loss, we adopt the approach from vanilla 3DGS \cite{3DGS_2023}, which combines L1 loss and D-SSIM loss to optimize the 3D Gaussian parameters:
\begin{equation}
	\mathcal{L}_r= (1 - \lambda_1)\mathcal{L}_1+ \lambda_1 \mathcal{L}_{\text{D-SSIM}}, 
\end{equation}
where $\lambda_1=0.2$ is the respective weight to balance the contributions from the L1 loss and D-SSIM loss.

For the medium model, precise depth information is crucial for accurately representing underwater scenes. We utilize DepthAnythingV2 \cite{Depth_Anything_V2}, a robust monocular depth estimation model, to generate pseudo ground truth depth maps $D'$. We transform our depth map to its inverse form $\mathcal{D} = 1/(D + 1)$, and compute the L1 loss between rendered and pseudo depths.

While the pseudo depth provides overall depth structure, direct rendering of depth using Eq. \eqref{eq: depth} can lead to artifacts, particularly at object boundaries where depth discontinuities naturally occur. The discrete nature of 3D Gaussians and their finite extent can introduce spurious depth variations that compromise the quality of novel view synthesis.
To address this issue while preserving legitimate depth discontinuities, we introduce an edge-aware depth smoothness regularization term that adapts its strength based on local image content:
\begin{equation}
    \mathcal{L}_{DS} = \frac{1}{P} \sum_{p=1}^{P} \left| \frac{\nabla D_p}{\max \left( \nabla I_p, \varepsilon \right)} \right|,
\end{equation}
where $\nabla$ represents the first-order spatial derivative operator encompassing both horizontal ($\nabla_x$) and vertical ($\nabla_y$) gradients, $I$ denotes the rendered image (treated as a constant during backpropagation), $P = H \times W$ represents the total number of pixels, and $\varepsilon$ is a small regularization constant to prevent numerical instability from division by zero.
This regularization uses image gradients to guide depth changes, enforcing smoothness only in regions where the image intensity changes significantly (i.e., edges or texture areas), while allowing more flexibility in depth changes in flatter regions of the image.

Additionally, in regions far from the camera’s line of sight, foreground and background objects may appear similar in color despite having significantly different depths.
In such cases, the smoothness term may over-regularize depth variations at true boundaries, resulting in the loss of important geometric details.
To complement the edge-aware term and ensure robust boundary preservation, we incorporate anisotropic total variation (TV) regularization:
\begin{equation}
\mathcal{L}_{DC} = \text{mean}(|\nabla_x D|) + \text{mean}(|\nabla_y D|),
\end{equation}
where $\nabla_x$ and $\nabla_y$ are spatial gradients in the directions of $x$ and $y$, respectively.

By summarizing all depth-related supervision and regularization terms, we define the complete depth loss function as:
\begin{equation}
	\mathcal{L}_{d} = \lambda_2 \|\mathcal{D} - D'\|_1 + \lambda_3\mathcal{L}_{DS} + \lambda_4 \mathcal{L}_{DC},
\end{equation}
where we set $\lambda_2 = 0.1$, $\lambda_3 = 0.01$, and $\lambda_4 =0.1 $. 

In underwater scene reconstruction, a volumetric semi-transparent water medium complicates accurate surface modeling. General 3D GS generates Gaussian centers $p_i$ positioned throughout these translucent volumes rather than adhering to solid surfaces, making them unsuitable for direct surface reconstruction in underwater environments.

We apply scale regularization to refine the 3D Gaussian ellipsoids into highly flat shapes, suppressing spurious floating artifacts caused by medium-object ambiguities \cite{PGSR}. This process narrows the Gaussians significantly, pulling their centers closer to actual physical surfaces.
Specifically, we penalize the minimal scaling component of each Gaussian:
\begin{equation}
	\mathcal{L}_s = \lambda_5 \sum_{i} \min(s_1^i, s_2^i, s_3^i),
\end{equation}
where $\lambda_5 =100$, and $s_{1}$, $s_{2}$, $s_{3}$ denote the activated scaling parameters of each Gaussian ellipsoid (i.e., $s_k = \exp(s_k^{\text{raw}})$, which are strictly positive by construction). Since the scaling values are always non-negative after exponential activation, the L1 norm reduces to a direct summation. Minimizing $\mathcal{L}_s$ effectively collapses the shortest axis of each ellipsoid, compressing Gaussians into surface-tangent discs that concentrate near physical surfaces, enabling clear disambiguation between the explicitly modeled water medium (via SMM) and solid object surfaces.
While this selective flattening strategy shares conceptual similarities with the explicit 2D disk primitives in \cite{2dgs}, we retain the 3D Gaussian framework to seamlessly integrate our explicit medium parameterization through the UIFM-based volumetric rendering pipeline.

\begin{figure*}[!h]
	\centering
	\includegraphics[width=1\linewidth]{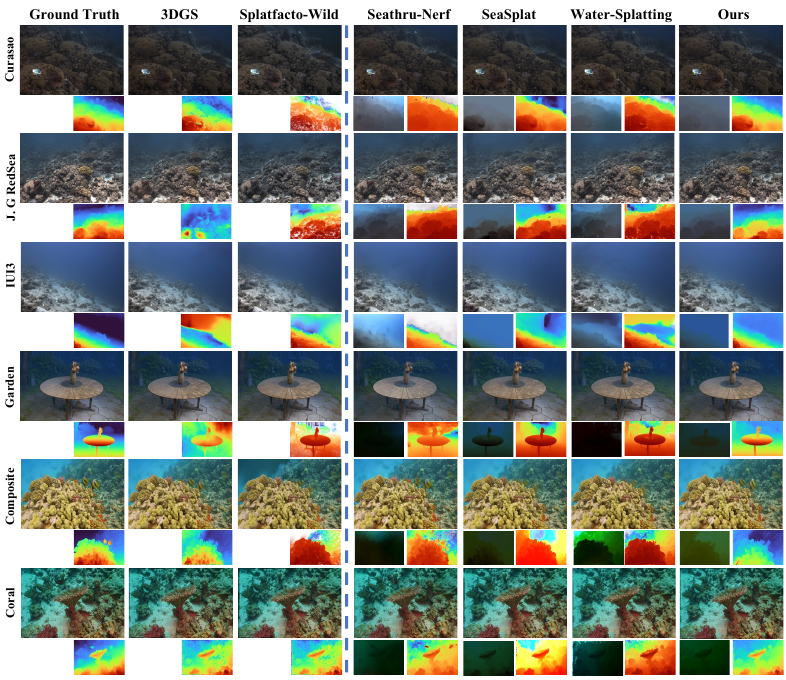} 
	% \vspace{-0.3in}
	\caption{Qualitative comparison of underwater NVS tasks on real-world and simulated datasets. We visualize the rendered depth maps and backscatter maps. The proposed method outperforms other approaches in capturing scene details and maintaining scene consistency, whereas other methods exhibit floating artifacts and inaccurate depth estimation (\emph{Zoom-in for best view}).} 
	\label{fig:all_result}
\end{figure*}

\section{Experiments}

%-------------------------------------------------------------------------
\subsection{Experimental Settings}
\noindent
\textbf{Datasets.} We evaluate our method on three datasets: SeaThru-NeRF~\cite{SeaThru-NeRF_2023_CVPR}, Underwater in the Wild (U-IW)~\cite{UNeRF_2024_CVPR}, and our Simulated Underwater dataset (S-U). SeaThru-NeRF contains four forward-facing real underwater scenes with diverse aquatic and imaging conditions. It encapsulates a wide range of water conditions and imaging scenarios, with unbounded camera-to-scene distances.
U-IW comprises densely sampled frames from in-the-wild underwater videos collected from the Internet. The dataset includes more complete camera trajectories, often covering 360° views around underwater scenes. This provides richer temporal and spatial cues, enabling more comprehensive evaluations under general real-world conditions.
To further evaluate our method, we simulated underwater scenes using four scenes from the MipNeRF-360 dataset \cite{Barron_2022_CVPR}. The construction follows the setup described in \cite{SeaThru-NeRF_2023_CVPR}. 
The water range is determined from depth maps predicted by 3DGS with Gsplat~\cite{gsplat_2024}.  Following the setup in \cite{SeaThru-NeRF_2023_CVPR}, underwater scenarios are simulated according to Eq. \eqref{eq:UIFM} with parameter values $\beta^D = [1.3, 1.2, 0.9]$, $\beta^B = [0.95, 0.85, 0.7]$, and $B^\infty = [0.07, 0.2, 0.39]$.
All scenes are preprocessed with COLMAP~\cite{SFM_2016_CVPR} to extract camera poses and initialize Gaussian positions from sparse point clouds.

\noindent
\textbf{Implementations.}
We adopt hyperparameter values from \cite{3DGS_2023} and train all scenes for 30k iterations. The scene modeling with one MLP contains three linear layers with 128 hidden units. We adopt the densification strategy of AbsGS \cite{absgs_acmmm_2024}. We employ the Adam optimizer with a learning rate of $10^{-3}$ for MLP training. All experiments are conducted on an NVIDIA RTX 3090 GPU.

\noindent
\textbf{Baselines and Metrics.}
We compare our method with several representative baselines retrained under the same setting, including vanilla 3DGS \cite{3DGS_2023}, Splatfacto-W (in the wild) \cite{Splatfacto-W_2024}, and recent underwater NVS methods such as SeaThru-NeRF  \cite{Sea-Thru_2019_CVPR} (implemented on Nerfstudio~\cite{nerfstudio}), RecGS \cite{RecGS_2024}, SeaSplat \cite{SeaSplat_2024}, and Water-Splatting \cite{watersplatting_2024}.
For each baseline method, we assess rendering quality using PSNR, SSIM, and LPIPS. All baseline methods are retrained under identical configurations using original dataset resolutions.

\begin{table*}[!]
	\centering
	\small 
	\definecolor{rank1}{RGB}{255,200,200}
	\definecolor{rank2}{RGB}{255,228,180}
	\definecolor{rank3}{RGB}{255,255,204}
	\caption{Quantitative comparison across datasets (averaged over all scenes).
		The top three rankings in each category are highlighted: \colorbox[rgb]{1.0, 0.784, 0.784}{1st}, \colorbox[rgb]{1.0, 0.894, 0.706}{2nd}, and \colorbox[rgb]{1.0, 1.0, 0.8}{3rd}. Arrows indicate whether higher (↑) or lower (↓) values are better.}
	\definecolor{rank1}{RGB}{255,200,200}
	\definecolor{rank2}{RGB}{255,228,180}
	\definecolor{rank3}{RGB}{255,255,204}
	\resizebox{1\linewidth}{!}{%
		\begin{tabular}{l%
				*{3}{c}%
				*{3}{c}%
				*{3}{c}%
				*{2}{c}}
			\toprule
			\multirow{2}{*} & \multicolumn{3}{c}{SeaThru-NeRF} & \multicolumn{3}{c}{U-IW} & \multicolumn{3}{c}{S-U} & \multicolumn{2}{c}{Speed} \\
			\cmidrule(lr){2-4} \cmidrule(lr){5-7} \cmidrule(lr){8-10} \cmidrule(lr){11-12}
			& PSNR ↑ & SSIM ↑ & LPIPS ↓ & PSNR↑ & SSIM ↑ & LPIPS ↓ & PSNR ↑ & SSIM ↑ & LPIPS ↓ & FPS ↑ & Training Time ↓ \\
			\midrule
			
			SeaThru-NeRF \cite{SeaThru-NeRF_2023_CVPR} & \cellcolor{rank3}27.394  & 0.860  & 0.215  & 18.942  & 0.644  & 0.383  & 24.436  & 0.805  & 0.293  & 0.55  & 2 h 39 m \\
			3DGS \cite{3DGS_2023} & 26.188  & 0.859  & 0.238  & \cellcolor{rank2}27.361  & \cellcolor{rank2}0.894  & \cellcolor{rank2}0.158  & \cellcolor{rank3}29.274  & \cellcolor{rank2}0.881  & \cellcolor{rank2}0.233  & 149.36  & 17 m \\
			Splatfacto-Wild \cite{Splatfacto-W_2024} & 25.750  & 0.832  & 0.229  & 25.159  & 0.847  & 0.209  & 25.786  & 0.853  & 0.260  & 42.98 & 21 m \\
			SeaSplat \cite{SeaSplat_2024}& 27.385  & \cellcolor{rank2}0.866  & \cellcolor{rank1}0.194  & \cellcolor{rank3}27.023  & \cellcolor{rank3}0.889  & \cellcolor{rank3}0.159  & 28.566  & 0.861  & 0.253  & 42.69 & 1 h 25 m \\
			RecGS \cite{RecGS_2024} & 25.829  & 0.857  & 0.233  & 22.186  & 0.838  & 0.180  & 24.620  & 0.825  & 0.259  & 146.62 & 38 m \\
			Water-Splatting \cite{watersplatting_2024} & \cellcolor{rank2}27.573  & \cellcolor{rank3}0.865  & \cellcolor{rank2}0.198  & 25.673  & 0.882  & 0.167  & \cellcolor{rank2}29.973  & \cellcolor{rank3}0.878  & \cellcolor{rank3}0.235  & 35.80 & 29 m \\
			Ours & \cellcolor{rank1}28.116  & \cellcolor{rank1}0.876  & \cellcolor{rank3}0.202  & \cellcolor{rank1}28.198  & \cellcolor{rank1}0.902  & \cellcolor{rank1}0.150  & \cellcolor{rank1}31.227  & \cellcolor{rank1}0.891  & \cellcolor{rank1}0.187  & 48.72 & 48 m \\
			
			\bottomrule
		\end{tabular}%
	}
	\label{tab:comparison}
\end{table*}

\begin{figure*}[!h]
	\centering
	\includegraphics[width=1\linewidth]{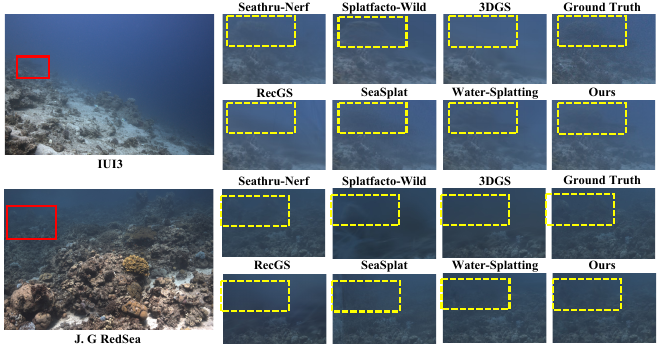} 
	% \vspace{-0.3in}
	\caption{Detailed visualization on SeaThru-NeRF dataset (IUI3 and J. G RedSea scenes). 
Zoomed-in regions (red boxes) demonstrate our superior detail preservation compared 
to baseline methods.} 
	\label{fig:detail_result}
\end{figure*}

%-------------------------------------------------------------------------
\subsection{Experimental Results}

\noindent
\textbf{Reconstruction Quality.} Figs.~\ref{fig:all_result} and \ref{fig:detail_result}
suggest that our method achieves superior reconstruction quality by effectively mitigating the effects of water-column artifacts and providing coherent depth estimation. 
In particular, for the geometry reconstruction, the enhanced detail recovery is most noticeable, which mitigates the effects of water-specific artifacts that plague underwater imaging.
While alternative approaches exhibit obvious floating artifacts and inaccurate distant scene reconstruction, our method maintains coherent scene structure throughout the water column. This is particularly evident in the Coral and Composite scenes, where complex underwater structures are preserved with significantly higher fidelity.
Furthermore, our approach ensures a more accurate and consistent depth map, where other methods exhibit floating artifacts and collapsed depth issues, especially in complex underwater environments. The inconsistent depth estimations and blurry scene details are caused by water-specific artifacts, which often hinder accurate underwater reconstruction in other methods.
The visual consistency observed across all tested datasets further demonstrates the robustness of our method in real-world underwater scenarios, where water conditions and scene complexities vary greatly. This consistency is particularly important when dealing with diverse aquatic environments, as our approach adapts to different underwater conditions with minimal loss in reconstruction quality.
Overall, the results show that our method excels in both scene restoration quality and depth estimation, leading to more realistic underwater scene representations and accurate scene geometry.

\begin{table*}[h]
	\centering
	\scriptsize 
	\setlength{\tabcolsep}{2.5pt} 
	\renewcommand{\arraystretch}{1} 
	
	\caption{Quantitative results: Results on four real-world underwater scenes from the SeaThru-NeRF dataset \cite{SeaThru-NeRF_2023_CVPR}: Curascao, IUI3-RedSea, J. G RedSea, and Panama. Evaluation metrics (PSNR/SSIM/LPIPS) demonstrate the effectiveness of different scene reconstruction methods, with top three rankings in each category highlighted:  \colorbox[rgb]{1.0, 0.784, 0.784}{1st}, \colorbox[rgb]{1.0, 0.894, 0.706}{2nd}, and \colorbox[rgb]{1.0, 1.0, 0.8}{3rd}.
	}
	\definecolor{rank1}{RGB}{255,200,200}
	\definecolor{rank2}{RGB}{255,228,180}
	\definecolor{rank3}{RGB}{255,255,204}
	\begin{adjustbox}{max width=\textwidth, scale=1.0} 
		\begin{tabular}{@{}l*{4}{ccc}@{}}
			\toprule
			& 
			\multicolumn{3}{c}{Curascao} & 
			\multicolumn{3}{c}{IUI3-RedSea} & 
			\multicolumn{3}{c}{J. G RedSea} & 
			\multicolumn{3}{c}{Panama} \\
			\cmidrule(lr){2-4} \cmidrule(lr){5-7} \cmidrule(lr){8-10} \cmidrule(lr){11-13}
			& PSNR (rank) & SSIM (rank) & LPIPS (rank) & PSNR (rank) & SSIM (rank) & LPIPS (rank) & PSNR (rank) & SSIM (rank) & LPIPS (rank) & PSNR (rank) & SSIM (rank) & LPIPS (rank) \\
			\midrule
			
			SeaThru-NeRF \cite{SeaThru-NeRF_2023_CVPR}
			& \cellcolor{rank2}30.870\ (2) & 0.891\ (6) & 0.222\ (7) 
			& 26.748\ (4) & 0.824\ (5) & 0.281\ (4) 
			& \cellcolor{rank3}23.213\ (3) & \cellcolor{rank3}0.844\ (3) & \cellcolor{rank3}0.193\ (3) 
			& 28.747\ (4) & 0.881\ (4) & \cellcolor{rank1}0.162\ (1) \\
			
			3DGS  \cite{3DGS_2023}
			& 29.601\ (6) & 0.893\ (4) & 0.220\ (6) 
			& 24.912\ (6) & 0.826\ (4) & 0.290\ (6) 
			& 21.316\ (7) & 0.835\ (5) & 0.223\ (6) 
			& \cellcolor{rank2}28.921\ (2) & \cellcolor{rank3}0.884\ (3) & 0.217\ (7) \\
			
			Splatfacto-Wild  \cite{Splatfacto-W_2024}
			& 28.048\ (7) & 0.876\ (7) & 0.209\ (4) 
			& 25.697\ (5) & 0.806\ (7) & 0.292\ (7) 
			& 22.962\ (4) & 0.815\ (7) & 0.204\ (5) 
			& 26.291\ (7) & 0.829\ (6) & 0.210\ (6) \\
			
			SeaSplat  \cite{SeaSplat_2024}
			& \cellcolor{rank3}30.768\ (3) & \cellcolor{rank3}0.898\ (3) & \cellcolor{rank2}0.186\ (2) 
			& \cellcolor{rank3}27.493\ (3) & \cellcolor{rank3}0.832\ (2) & \cellcolor{rank1}0.213\ (1) 
			& 22.698\ (5) & \cellcolor{rank2}0.845\ (2) & 0.196\ (4) 
			& 28.580\ (5) & \cellcolor{rank2}0.887\ (2) & 0.180\ (4) \\
			
			RecGS  \cite{RecGS_2024}
			& 29.313\ (6) & 0.893\ (4) & 0.211\ (5) 
			& 24.363\ (7) & 0.823\ (6) & 0.289\ (5) 
			& 21.812\ (6) & 0.834\ (6) & 0.224\ (7) 
			& 27.829\ (6) & 0.879\ (7) & 0.208\ (5) \\
			
			Water-Splatting  \cite{watersplatting_2024}
			& 30.583\ (4) & \cellcolor{rank2}0.904\ (2) & \cellcolor{rank1}0.174\ (1) 
			& \cellcolor{rank2}27.593\ (2) & \cellcolor{rank3}0.831\ (3) & \cellcolor{rank3}0.275\ (3) 
			& \cellcolor{rank2}23.363\ (2) & \cellcolor{rank2}0.845\ (2) & \cellcolor{rank1}0.178\ (1) 
			& \cellcolor{rank3}28.751\ (3) & 0.881\ (4) & \cellcolor{rank2}0.163\ (2) \\
			
			Ours
			& \cellcolor{rank1}30.978\ (1) & \cellcolor{rank1}0.907\ (1) & \cellcolor{rank3}0.187\ (3) 
			& \cellcolor{rank1}28.301\ (1) & \cellcolor{rank1}0.841\ (1) & \cellcolor{rank2}0.252\ (2) 
			& \cellcolor{rank1}23.371\ (1) & \cellcolor{rank1}0.857\ (1) & \cellcolor{rank2}0.187\ (2) 
			& \cellcolor{rank1}29.816\ (1) & \cellcolor{rank1}0.900\ (1) & \cellcolor{rank3}0.181\ (3) \\
			\bottomrule
		\end{tabular}
	\end{adjustbox}
	\label{tab:comparison1}
\end{table*}

\begin{table*}[!h]
	\centering
	\scriptsize
	\setlength{\tabcolsep}{4.2pt}
	\renewcommand{\arraystretch}{1}
	\caption{
		Quantitative results: Results on four in-the-wild underwater scenes from the U-IW dataset \cite{UNeRF_2024_CVPR}:  Turtle, Coral, Sardine, and Composite. Evaluation metrics (PSNR/SSIM/LPIPS) demonstrate the effectiveness of different scene reconstruction methods, with top three rankings in each category highlighted: \colorbox[rgb]{1.0, 0.784, 0.784}{1st}, \colorbox[rgb]{1.0, 0.894, 0.706}{2nd}, and \colorbox[rgb]{1.0, 1.0, 0.8}{3rd}.}
	
	\definecolor{rank1}{RGB}{255,200,200}
	\definecolor{rank2}{RGB}{255,228,180}
	\definecolor{rank3}{RGB}{255,255,204}
	\begin{adjustbox}{max width=\textwidth, scale=1.0} 
		\begin{tabular}{@{}l*{4}{ccc}@{}}
			\toprule
			& 
			\multicolumn{3}{c}{Turtle} & 
			\multicolumn{3}{c}{Coral} & 
			\multicolumn{3}{c}{Sardine} & 
			\multicolumn{3}{c}{Composite} \\
			\cmidrule(lr){2-4} \cmidrule(lr){5-7} \cmidrule(lr){8-10} \cmidrule(lr){11-13}
			& PSNR (rank) & SSIM (rank) & LPIPS (rank) & PSNR (rank) & SSIM (rank) & LPIPS (rank) & PSNR (rank) & SSIM (rank) & LPIPS (rank) & PSNR (rank) & SSIM (rank) & LPIPS (rank) \\
			\midrule		
			SeaThru-NeRF  \cite{SeaThru-NeRF_2023_CVPR}
			& 13.067\ (7) & 0.542\ (7) & 0.502\ (8) 
			& 25.354\ (7) & 0.855\ (7) & 0.160\ (6) 
			& 13.893\ (7) & 0.383\ (7) & 0.624\ (7) 
			& 23.456\ (6) & 0.795\ (7) & 0.248\ (7) \\
			
			3DGS \cite{3DGS_2023}
			& \cellcolor{rank3}32.145\ (3) & \cellcolor{rank2}0.958\ (2) & \cellcolor{rank2}0.112\ (2) 
			& \cellcolor{rank2}28.432\ (2) & \cellcolor{rank1}0.907\ (1) & \cellcolor{rank2}0.114\ (2) 
			& \cellcolor{rank2}22.797\ (2) & 0.838\ (4) & 0.247\ (4) 
			& \cellcolor{rank3}26.071\ (3) & \cellcolor{rank2}0.874\ (2) & \cellcolor{rank3}0.158\ (3) \\
			
			Splatfacto-Wild \cite{Splatfacto-W_2024}
			& 28.324\ (4) & 0.929\ (4) & 0.158\ (5) 
			& 25.788\ (6) & 0.868\ (6) & 0.162\ (7) 
			& 19.941\ (5) & 0.766\ (5) & 0.286\ (5) 
			& \cellcolor{rank2}26.585\ (2) & 0.826\ (6) & 0.232\ (6) \\

			SeaSplat \cite{SeaSplat_2024}
			& \cellcolor{rank2}32.294\ (2) & \cellcolor{rank3}0.957\ (3) & \cellcolor{rank1}0.107\ (1) 
			& 27.595\ (4) & 0.895\ (5) & 0.124\ (5) 
			& \cellcolor{rank3}22.703\ (3) & \cellcolor{rank3}0.839\ (3) & \cellcolor{rank3}0.241\ (3) 
			& 25.500\ (5) & 0.864\ (5) & 0.165\ (5) \\
			
			RecGS \cite{RecGS_2024}
			& 24.836\ (6) & 0.875\ (6) & \cellcolor{rank3}0.116\ (3) 
			& 25.970\ (5) & \cellcolor{rank3}0.904\ (3) & 0.115\ (4) 
			& 15.343\ (6) & 0.708\ (6) & 0.330\ (6) 
			& 22.596\ (7) & 0.866\ (4) & 0.162\ (4) \\
			
			Water-Splatting \cite{watersplatting_2024}
			& 26.075\ (5) & 0.912\ (5) & 0.174\ (6) 
			& \cellcolor{rank3}28.225\ (3) & 0.902\ (4) & \cellcolor{rank1}0.113\ (1) 
			& 22.666\ (4) & \cellcolor{rank2}0.843\ (2) & \cellcolor{rank2}0.237\ (2) 
			& 25.727\ (4) & \cellcolor{rank3}0.873\ (3) & \cellcolor{rank1}0.146\ (1) \\
			
			Ours
			& \cellcolor{rank1}32.538\ (1) & \cellcolor{rank1}0.959\ (1) & 0.117\ (4) 
			& \cellcolor{rank1}28.735\ (1) & \cellcolor{rank1}0.907\ (1) & \cellcolor{rank2}0.114\ (2) 
			& \cellcolor{rank1}24.974\ (1) & \cellcolor{rank1}0.867\ (1) & \cellcolor{rank1}0.218\ (1) 
			& \cellcolor{rank2}26.546\ (2) & \cellcolor{rank1}0.877\ (1) & \cellcolor{rank2}0.153\ (2) \\
			\bottomrule
		\end{tabular}
	\end{adjustbox}
	\label{tab:comparison2}
\end{table*}

\begin{table*}[!]
	\centering
	\scriptsize
	\setlength{\tabcolsep}{4.2pt}
	\renewcommand{\arraystretch}{1}
	\caption{
		Quantitative results: Results on four synthetic 360 scenes from the S-U dataset \cite{Barron_2022_CVPR}: Garden, Bicycle, Stump, and Counter. Evaluation metrics (PSNR/SSIM/LPIPS) demonstrate the effectiveness of different scene reconstruction methods, with top three rankings in each category highlighted:  \colorbox[rgb]{1.0, 0.784, 0.784}{1st}, \colorbox[rgb]{1.0, 0.894, 0.706}{2nd}, and \colorbox[rgb]{1.0, 1.0, 0.8}{3rd}.}
	\definecolor{rank1}{RGB}{255,200,200}
	\definecolor{rank2}{RGB}{255,228,180}
	\definecolor{rank3}{RGB}{255,255,204}
	\begin{adjustbox}{max width=\textwidth, scale=1.0} 
		\begin{tabular}{@{}l*{4}{ccc}@{}}
			\toprule
			& 
			\multicolumn{3}{c}{Garden} & 
			\multicolumn{3}{c}{Bicycle} & 
			\multicolumn{3}{c}{Stump} & 
			\multicolumn{3}{c}{Counter} \\
			\cmidrule(lr){2-4} \cmidrule(lr){5-7} \cmidrule(lr){8-10} \cmidrule(lr){11-13}
			& PSNR (rank) & SSIM (rank) & LPIPS (rank) & PSNR (rank) & SSIM (rank) & LPIPS (rank) & PSNR (rank) & SSIM (rank) & LPIPS (rank) & PSNR (rank) & SSIM (rank) & LPIPS (rank) \\
			\midrule
			
			SeaThru-NeRF \cite{SeaThru-NeRF_2023_CVPR}
			& 26.024\ (7) & 0.884\ (7) & \cellcolor{rank3}0.188\ (3) 
			& 22.391\ (7) & 0.730\ (7) & 0.361\ (7) 
			& 25.110\ (5) & 0.738\ (6) & 0.364\ (7) 
			& 24.221\ (6) & 0.867\ (6) & 0.259\ (7) \\
			
			3DGS \cite{3DGS_2023}
			& \cellcolor{rank2}32.515\ (2) & \cellcolor{rank2}0.922\ (2) & 0.191\ (4) 
			& \cellcolor{rank3}28.136\ (3) & \cellcolor{rank2}0.856\ (2) & \cellcolor{rank3}0.242\ (3) 
			& 25.497\ (4) & \cellcolor{rank3}0.808\ (3) & \cellcolor{rank3}0.320\ (3) 
			& \cellcolor{rank2}30.947\ (2) & \cellcolor{rank2}0.937\ (2) & \cellcolor{rank2}0.178\ (2) \\
			
			Splatfacto-Wild \cite{Splatfacto-W_2024}
			& 27.140\ (6) & 0.894\ (6) & 0.224\ (7) 
			& 24.491\ (5) & 0.814\ (6) & 0.280\ (5) 
			& 22.747\ (6) & 0.774\ (5) & 0.335\ (5) 
			& 28.765\ (5) & \cellcolor{rank3}0.928\ (3) & \cellcolor{rank3}0.201\ (3) \\

			SeaSplat \cite{SeaSplat_2024}
			& 30.524\ (4) & 0.896\ (5) & 0.209\ (6) 
			& 27.896\ (4) & \cellcolor{rank3}0.836\ (3) & 0.258\ (4) 
			& \cellcolor{rank3}26.531\ (3) & 0.800\ (4) & \cellcolor{rank3}0.320\ (3) 
			& 29.312\ (4) & 0.911\ (5) & 0.225\ (5) \\
			
			RecGS \cite{RecGS_2024}
			& 27.416\ (5) & 0.899\ (4) & 0.207\ (5) 
			& 26.144\ (6) & 0.831\ (4) & \cellcolor{rank2}0.237\ (2) 
			& 22.649\ (7) & 0.745\ (7) & 0.335\ (5) 
			& 22.270\ (7) & 0.826\ (7) & 0.256\ (6) \\
			
			Water-Splatting \cite{watersplatting_2024}
			& \cellcolor{rank3}31.897\ (3) & \cellcolor{rank2}0.922\ (2) & \cellcolor{rank2}0.175\ (2) 
			& \cellcolor{rank2}28.381\ (2) & 0.829\ (5) & 0.280\ (5) 
			& \cellcolor{rank1}29.097\ (1) & \cellcolor{rank1}0.837\ (1) & \cellcolor{rank2}0.279\ (2) 
			& \cellcolor{rank3}30.517\ (3) & 0.925\ (4) & 0.208\ (4) \\
			
			Ours
			& \cellcolor{rank1}33.203\ (1) & \cellcolor{rank1}0.923\ (1) & \cellcolor{rank1}0.142\ (1) 
			& \cellcolor{rank1}30.253\ (1) & \cellcolor{rank1}0.875\ (1) & \cellcolor{rank1}0.181\ (1) 
			& \cellcolor{rank2}28.975\ (2) & \cellcolor{rank2}0.832\ (2) & \cellcolor{rank1}0.255\ (1) 
			& \cellcolor{rank1}32.475\ (1) & \cellcolor{rank1}0.934\ (1) & \cellcolor{rank1}0.171\ (1) \\
			\bottomrule
		\end{tabular}
	\end{adjustbox}
	\label{tab:comparison3}
\end{table*}

\begin{figure}[!t]
	\includegraphics[width=1.0\linewidth]{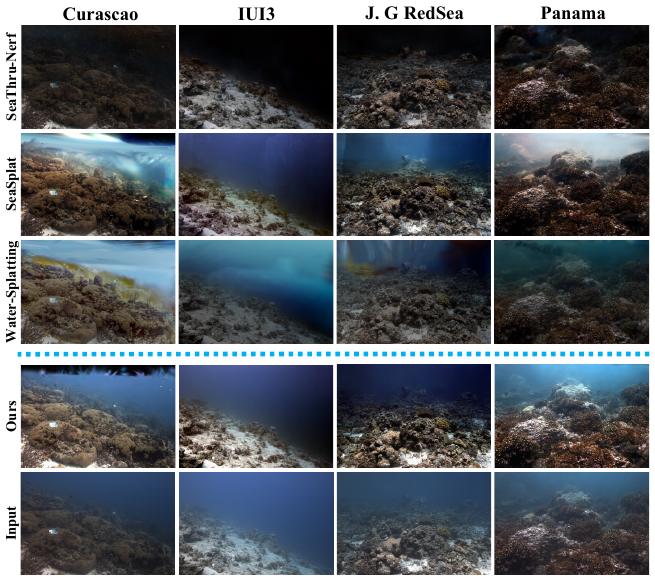}  
	\caption{Visualization of novel view restoration on the SeaThru-NeRF Dataset \cite{SeaThru-NeRF_2023_CVPR}. Our method achieves superior restoration quality with impressive visibility and scene consistency compared to other underwater NVS methods.}  
	\label{fig: restoration_seathru}
    \vspace{-2em}
\end{figure}

\begin{figure}[!t]
	\includegraphics[width=1.0\linewidth]{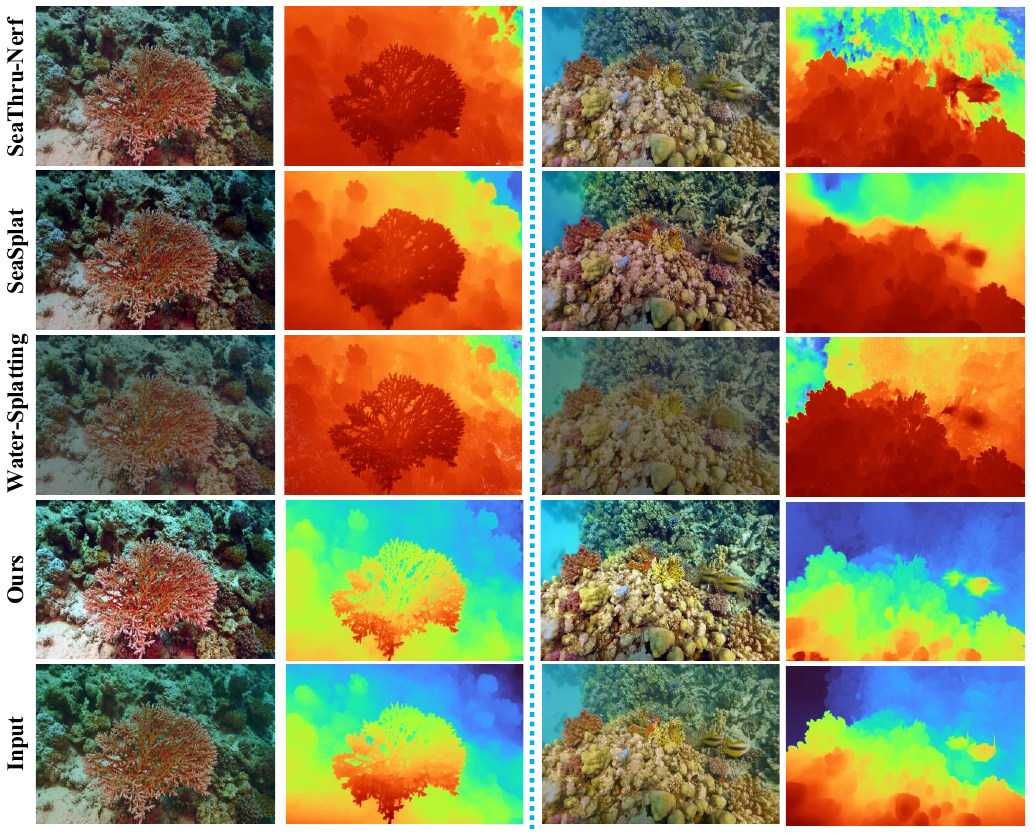}  
	\caption{Visualization of novel view restoration on the U-IW Dataset \cite{UNeRF_2024_CVPR}. Our method achieves superior restoration quality with accurate exposure balance across diverse scenes, outperforming other underwater NVS methods.}  
	\label{fig: restoration_uiw}
     \vspace{-1em}
\end{figure}

\begin{figure}[!]
	\includegraphics[width=1.0\linewidth]{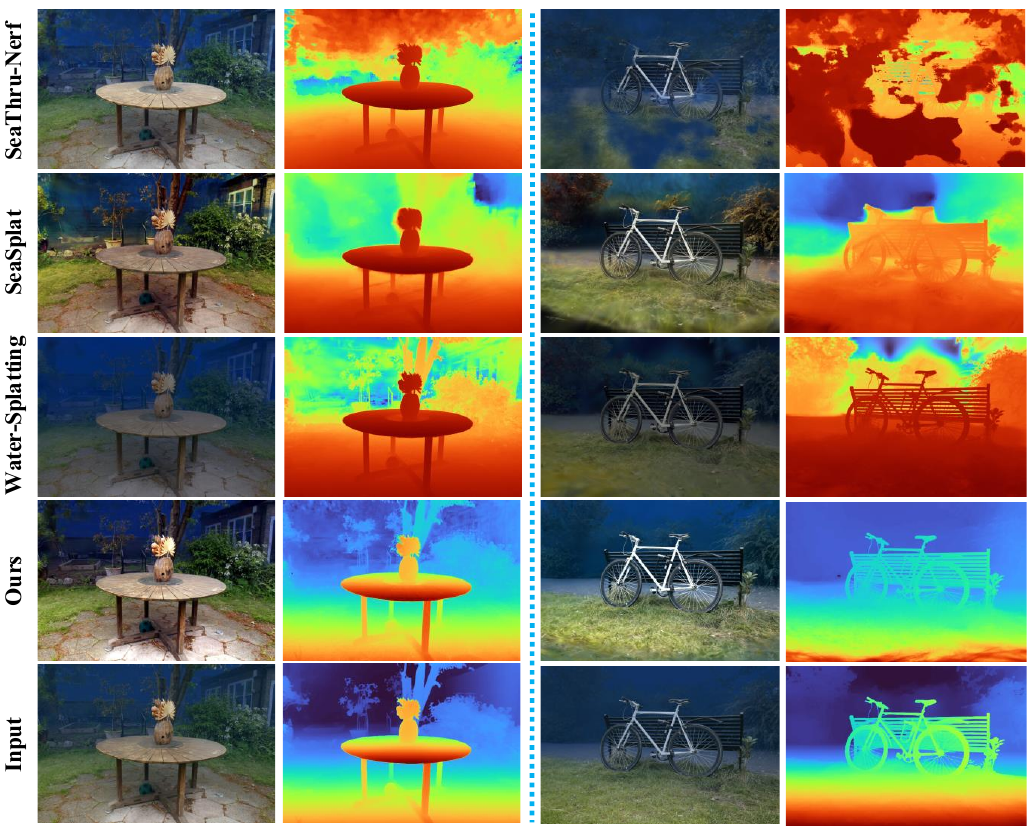}  
	\caption{Visualization of novel view restoration on the S-U Dataset. Our method achieves superior restoration quality, particularly excelling in the preservation of depth details compared to other underwater NVS methods.}  
	\label{fig: restoration_synthesis}
     \vspace{-2em}
\end{figure}

\noindent
\textbf{Scene Restoration.} Figs. \ref{fig: restoration_seathru}, \ref{fig: restoration_uiw}, and  \ref{fig: restoration_synthesis} present the scene restoration results for novel views. 
3DGS \cite{3DGS_2023}, Splatfacto-Wild \cite{Splatfacto-W_2024}, and RecGS \cite{RecGS_2024} lack medium modeling in their rendering pipelines, thus failing to achieve effective de-scattering restoration and are excluded from these comparisons. For these visual results of scene restoration, our method achieves superior quality by enhancing underwater visibility and preserving structural and temporal consistency more effectively than existing NVS methods.
The visual enhancements are most prominent in the accurate preservation of scene depth and structure, where our approach outperforms others in terms of global scene restoration.
Specifically, the proposed appearance modeling and depth-guided optimization contribute to more natural and coherent background restorations, especially in distant regions. The depth-aware representation ensures that both foreground and background elements are restored with appropriate spatial relationships, further enhancing the realism of the restored scene.
Additionally, the 2D white balance during the rendering process ensures a more natural appearance across varying illumination conditions and further enhances the visual realism of the restored views.

To quantitatively evaluate the restoration quality, we employ three established underwater image quality assessment (IQA) metrics: UCIQE (Underwater Color Image Quality Evaluation) \cite{uciqe}, UIQM (Underwater Image Quality Measure) \cite{uiqm}, and URanker \cite{uranker}. These metrics are specifically designed to assess underwater image restoration from complementary perspectives, including colorfulness, sharpness, contrast, and perceptual quality. Table \ref{tab:underwater_iqa} presents the averaged results across all scenes in the SeaThru-NeRF and U-IW datasets. Our method achieves the best overall performance, demonstrating substantial improvements over existing underwater NVS methods. These underwater-specific metrics validate that our physics-based appearance-medium decoupling produces perceptually superior restoration results that align with underwater image quality standards.

\begin{table}[t]
\centering
\caption{Underwater image quality metrics averaged across SeaThru-NeRF and U-IW datasets. $\uparrow$ indicates higher is better. Best results are \textbf{bold}.}
	\renewcommand{\arraystretch}{1.2}
     \resizebox{0.9\linewidth}{!}{
    \begin{tabular}{lccc}
        \toprule
        Method & UCIQE $\uparrow$ & UIQM $\uparrow$ & URanker $\uparrow$ \\
        \midrule
        SeaThru-NeRF \cite{SeaThru-NeRF_2023_CVPR} & 0.57 & 2.74 & 0.87 \\
        Water-Splatting \cite{watersplatting_2024} & 0.58 & 2.53 & 0.75 \\
        SeaSplat \cite{SeaSplat_2024} & 0.61 & 2.75 & 1.62 \\
        \midrule 
        \textbf{Ours}
        & \textbf{0.66} & \textbf{2.88} & \textbf{1.70} \\
        \bottomrule
    \end{tabular}}
\label{tab:underwater_iqa}
\end{table}

Furthermore, we provide an interactive comparison web demo on our \href{https://bilityniu.github.io/3D-UIR}{\textbf{project page}} for additional visual results.
This demo offers dynamic visual support for comprehensive cross-method evaluation, enabling real-time toggling between RGB renderings, depth estimations, and descattering-enhanced results across different approaches.

\noindent
\textbf{Quantitative Evaluations.}  
Table \ref{tab:comparison} presents the average metrics across all dataset scenes, while Tables \ref{tab:comparison1}, \ref{tab:comparison2}, and \ref{tab:comparison3} provide per-scene metric tables detailing quantitative evaluations across all test scenarios.
For quantitative results, our method consistently outperforms state-of-the-art approaches across three challenging datasets. We achieve the highest PSNR and SSIM values while maintaining low LPIPS scores on three datasets. 
The per-scene evaluation scores further reveal our method maintains this performance advantage consistently across diverse underwater scenes, from unbounded to surround scenes, and from simple to complex underwater scene geometries. 
Notably, our method performs well in challenging scenarios, even when optimal LPIPS scores are not achieved in certain scenes.
This trade-off highlights the inherent challenges in underwater scene reconstruction, where improvements in one visual quality aspect may sometimes compromise another. Nevertheless, our approach achieves the best balance across all metrics when considered holistically, prioritizing accurate scene reconstruction while maintaining competitive perceptual quality.
These comparison results highlight the effectiveness of our approach in handling underwater scene degradation and reconstructing high-fidelity visual content.

\noindent 
\textbf{Rendering Efficiency.} Table~\ref{tab:comparison} shows rendering speed and training time across four SeaThru-NeRF scenes. Our method achieves faster rendering and training than SeaThru-NeRF despite the latter's Nerfstudio acceleration. While our scene-medium modeling and depth optimization introduce slight computational overhead compared to vanilla 3DGS, we maintain real-time rendering with superior visual quality among underwater media methods, demonstrating an effective balance between reconstruction fidelity and efficiency.

\begin{figure}[!]
	\includegraphics[width=1.0\linewidth]{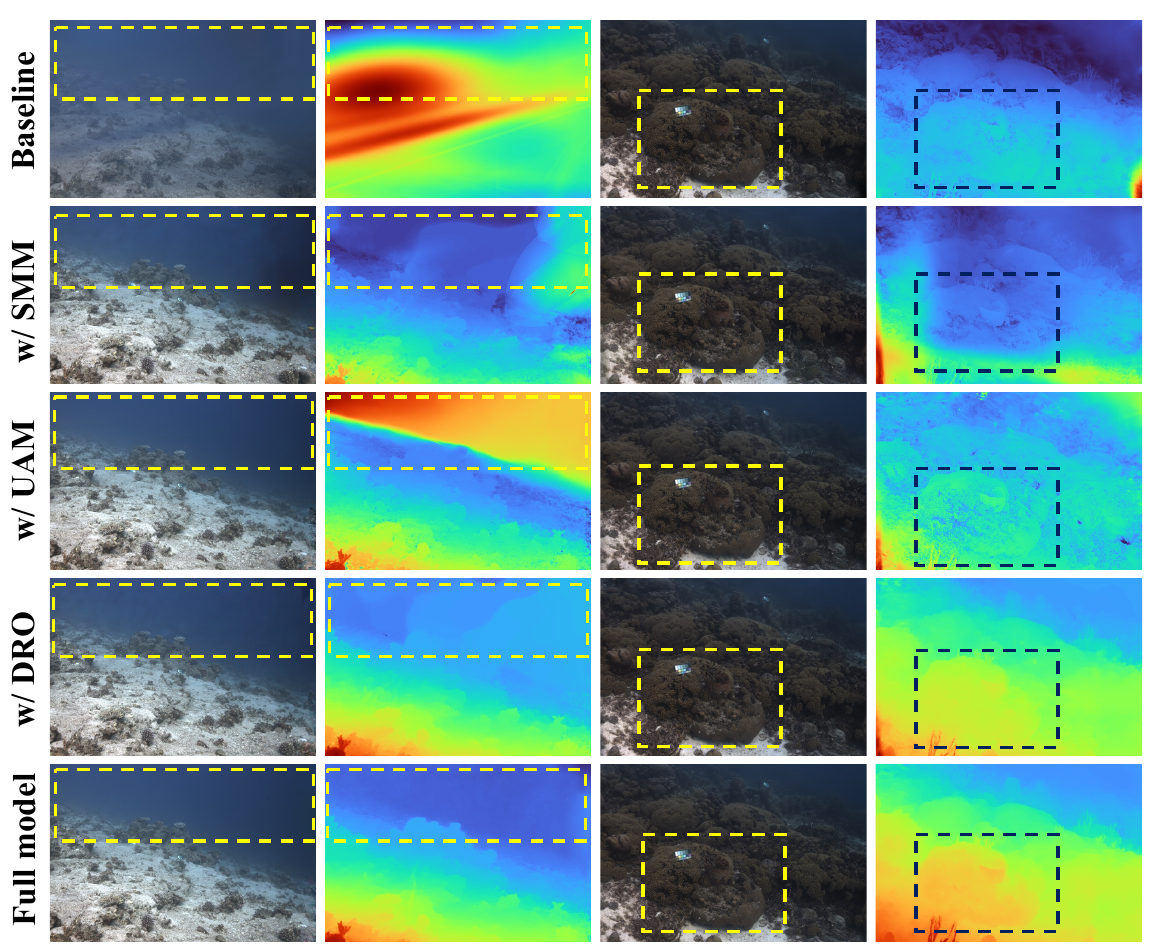} 
	\caption{Visualization of ablation studies on different components. Depth maps better illustrate medium modeling effects. Highlighted regions (dashed boxes) reveal progressive improvements in detail preservation and consistency across model variants. }  
	\label{fig: ablation}
     \vspace{-1em}
\end{figure}

%-------------------------------------------------------------------------
\subsection{Ablation Study}
\begin{figure}[!t]
	\includegraphics[width=1.0\linewidth]{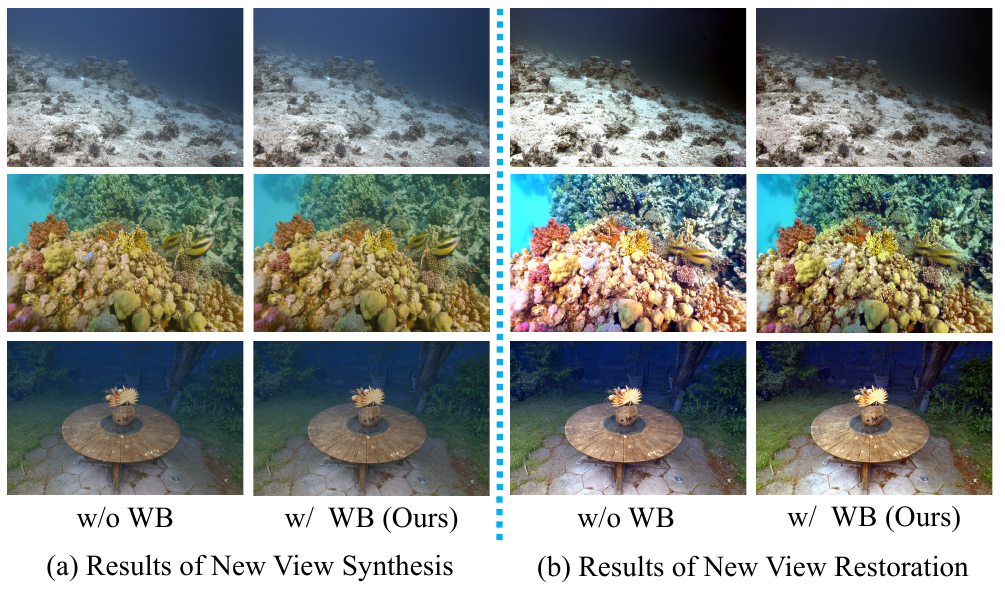} 
	\caption{
     Visualization of ablation studies on white balance. (a): novel view synthesis results demonstrate that pixel-based white balance does not disturb geometry reconstruction; (b): the corresponding restored views show that white balance achieves more visually balanced restorations by avoiding both over-enhancement and under-restoration.}  
	\label{fig: ablation_wb}
\end{figure}

% \noindent \textbf{Ablation Study.}
We conduct ablation studies on the SeaThru-NeRF datasets with results averaged across all scenes in Figs. \ref{fig: ablation}, \ref{fig: ablation_wb} and Table \ref{tab:ablation}.
Our experiments validate the effectiveness of three key modules. The scene disentanglement strategy shows significant improvements over the baseline 3DGS model when adding SMM and UAM separately. 
The medium model enhances accurate representation of underwater media by reducing floating artifacts caused by translucent medium occlusion while enabling scene restoration capabilities. Meanwhile, the appearance model improves color representation balance and achieves consistent brightness restoration across viewpoints.
Further performance gains are achieved when combining both components.
This synergistic integration demonstrates that explicit modeling of both intrinsic scene properties and water-induced effects is essential for accurate reconstruction.
Among optimization strategies, DRO yields substantial improvement, confirming that explicit depth supervision effectively mitigates volumetric artifacts in challenging underwater scenes. Various loss combinations provide additional refinement when integrated with the core components. 
Our full model achieves the best performance across all metrics, demonstrating the effectiveness of our comprehensive framework for underwater scene reconstruction.

\begin{figure}[t]
	\centering
	\includegraphics[width=1\linewidth]{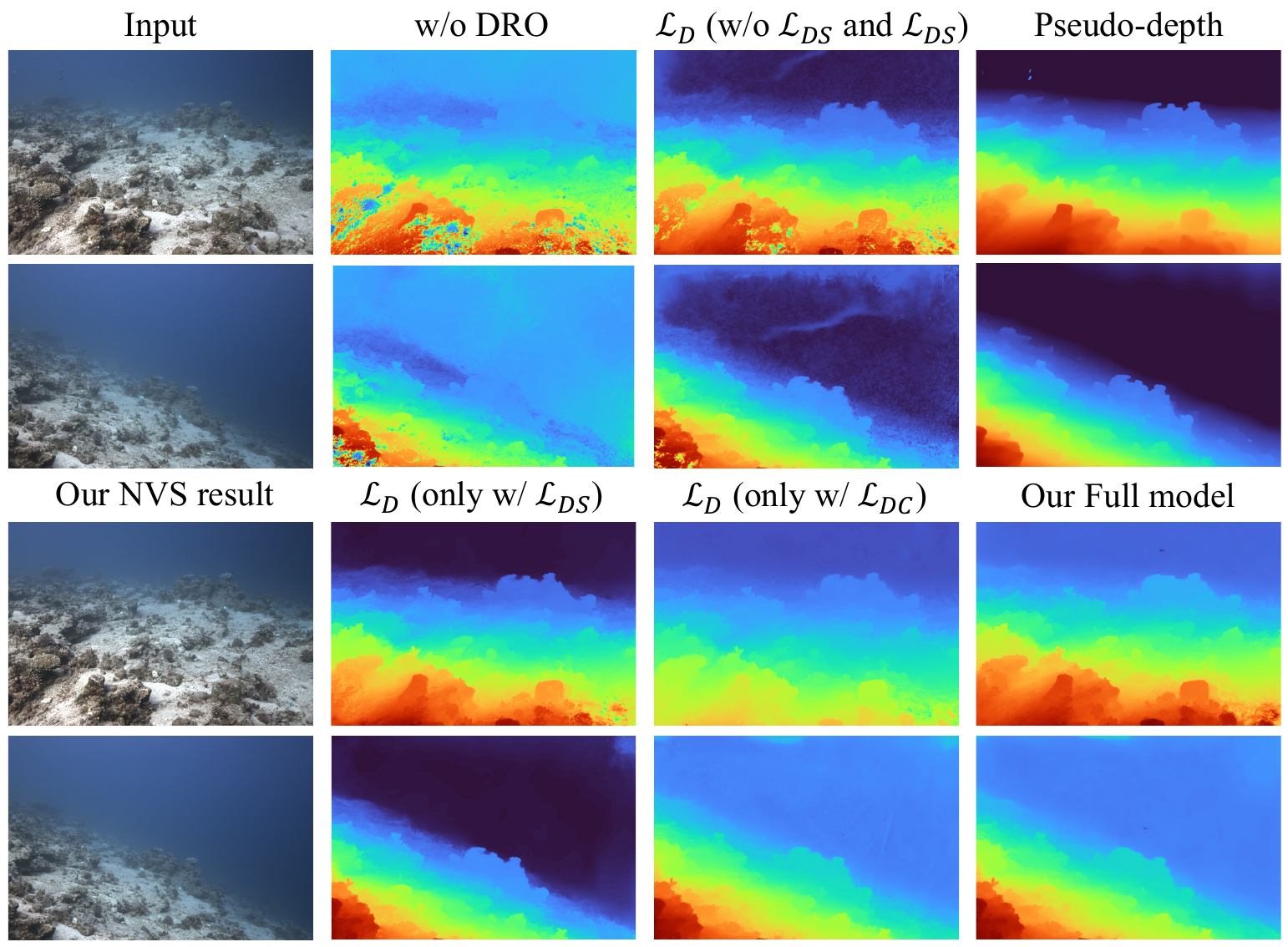}
	\caption{Ablation study on depth regularization components. We compare rendered depth maps and novel view synthesis results under different depth supervision strategies on two test views from the IUI3 scene.}
	\label{fig:depthloss}
\end{figure}

\begin{table}[t]  
	\centering 
	\caption{Ablation study results for evaluating the contribution of each component to the overall performance. Loss experiments are conducted on the SMM+UAM foundation. }
	\renewcommand{\arraystretch}{1}
     \resizebox{0.98\linewidth}{!}{
	\begin{tabular}{ll|c|c|c}    
		\toprule          Method &       & PSNR $\uparrow$  & SSIM $\uparrow$  & LPIPS $\downarrow$ \\    \midrule    
		\multicolumn{2}{l|}{Baseline} & 26.188 & 0.859 & 0.238 \\    \midrule   
		\multicolumn{2}{l|}{only w/ Medium (SMM)} & 26.698 & 0.864 & 0.216 \\    \midrule   
		\multicolumn{2}{l|}{only w/ Appearance (UAM)} & 27.245 & 0.867 & 0.210 \\    \midrule   
		\multicolumn{2}{l|}{only w/ Depth Optimization (DRO)} & 26.769 & 0.861 & 0.215 \\    \midrule
		\multicolumn{1}{c|}{SMM} & $\mathcal{L}_c$ & 27.507 & 0.868 & 0.209 \\\cmidrule{2-5}    
		\multicolumn{1}{c|}{$+$} & $\mathcal{L}_c + \mathcal{L}_d$ & 27.839 & 0.874 & 0.205 \\\cmidrule{2-5}    
		\multicolumn{1}{c|}{UAM} & $\mathcal{L}_c + \mathcal{L}_s$ & 27.691 & 0.867 & 0.211 \\    \midrule
        \multicolumn{2}{l|}{w/o Appearance Embedding in UAM} & 27.601 & 0.873 & 0.211 \\    \midrule   
        \multicolumn{2}{l|}{w/o Pixel based White Balance (ACS)} & 28.083 & 0.876 & 0.203 \\    \midrule   
		\multicolumn{2}{l|}{\textbf{Ours (Full Model)}} & \textbf{28.116} & \textbf{0.877} & \textbf{0.202} \\        \bottomrule
	\end{tabular}}%  
	\label{tab:ablation}%
\end{table}%

To further validate our physics-based modeling, we analyze how our framework handles imperfect depth priors. As shown in Fig. \ref{fig:depthloss}, our DRO combines $\mathcal{L}_{DS}$ and $\mathcal{L}_{DC}$ to achieve complementary effects. Using only $\mathcal{L}_{DS}$ tends to overfit the DepthAnything prior, while $\mathcal{L}_{DC}$ alone causes oversmoothing in near-field regions. Our dual-regularization design achieves an optimal balance by preserving structural boundaries while removing Gaussian-induced spike artifacts. 
This robustness is attributed to our physics-based optimization mechanism. In contrast to existing underwater NVS approaches that treat pseudo-depth as immutable ground truth, our method utilizes it primarily as structural initialization for optimization warm-start. Critically, the rendered depth serves as input to the SMM branch, establishing a differentiable physics-based closed loop. Inaccurate depth estimates propagate to erroneous medium parameter predictions ($\beta_D$, $\beta_B$, $B_\infty$), which subsequently manifest as photometric inconsistencies in the rendered imagery. These errors generate gradients that backpropagate through the differentiable pipeline to refine the underlying Gaussian geometry. Consequently, the physics-based rendering loss constrains the geometry to converge toward physically plausible configurations that accurately model the scattering phenomena, effectively correcting initial depth estimates through multi-view photometric consistency rather than rigidly overfitting to potentially erroneous pseudo-depth supervision.

Furthermore, we also analyze the restored water-free scenes to validate the effectiveness of the final pixel-based white balance. As shown in Fig. \ref{fig: ablation_wb}, incorporating pixel-based white balance during training leads to better restoration, avoiding over- and under-restoration. Additionally, it does not affect the critical geometric accuracy during GS training, as evidenced by the nearly identical visual results in Fig. \ref{fig: ablation_wb} and the closely matched metrics in Table \ref{tab:ablation}, which demonstrates its effectiveness in complex real-world scenarios.

\section{Conclusion}
This paper proposes a physics-based 3DGS framework that explicitly models underwater light transport by disentangling scene representation into object appearance and medium components. Combined with learned embeddings for view-dependent variations, depth-guided optimization, and Gaussian flattening constraints, our method effectively addresses floating artifacts and photometric inconsistencies caused by wavelength-dependent attenuation. Experiments demonstrate state-of-the-art novel view synthesis quality with real-time physically accurate scene restoration, expanding the possibilities for subsequent underwater view synthesis enhancement.

\small
\bibliographystyle{IEEEtran}
\bibliography{IEEEabrv,paper}

\vspace{-1.2cm}

\begin{IEEEbiography}
[{\includegraphics[width=1in,height=1.25in,clip,keepaspectratio]{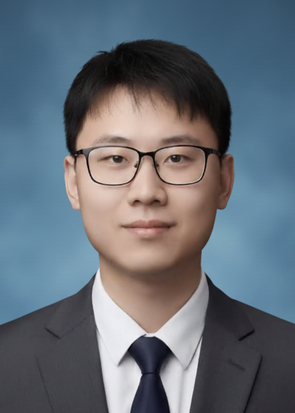}}]{Jieyu Yuan}
received the Ph.D. degree in computer technology and application from the School of Computer Science and Engineering, Faculty of Innovation Engineering, Macau University of Science and Technology, Macau, China, in 2024. He is currently a Post-Doctoral Researcher with the School of Computer Science, Nankai University, Tianjin, China. 

His research interests include computer vision, 3D reconstruction, and multimodal large models, with a particular focus on underwater vision.
\end{IEEEbiography}

\vspace{-1.3cm}
\begin{IEEEbiography}
[{\includegraphics[width=1in,height=1.25in,clip,keepaspectratio]{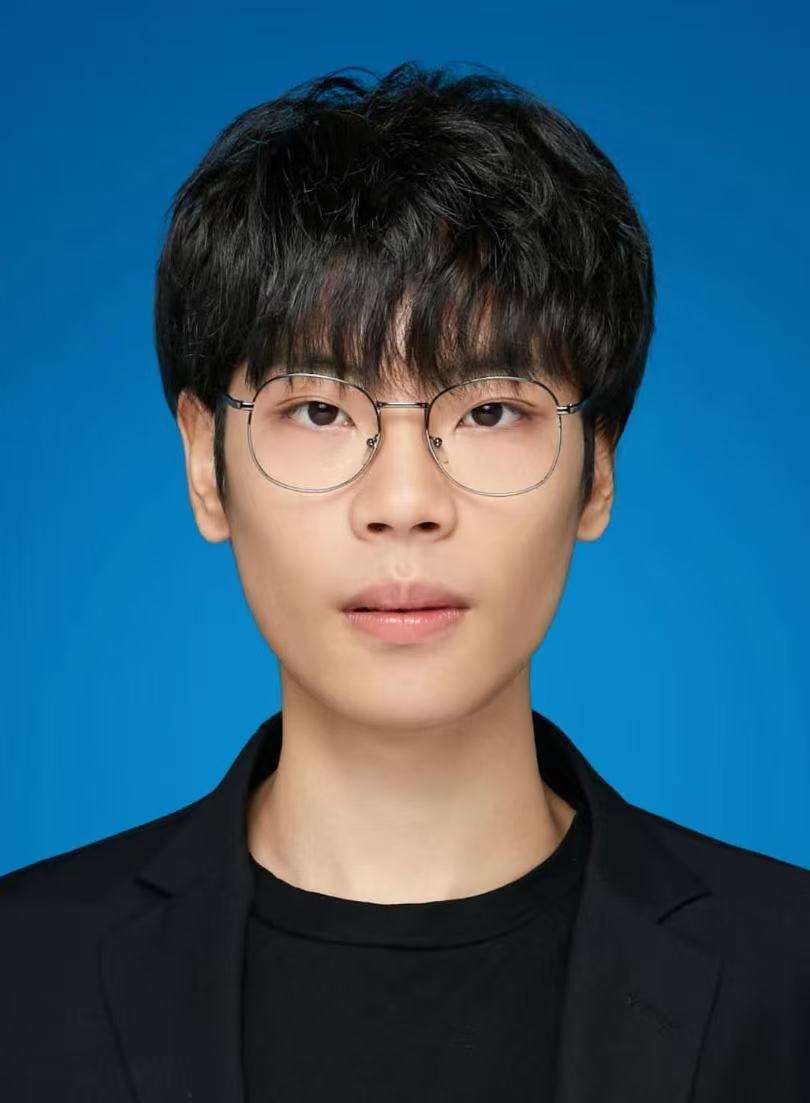}}]{Yujun Li}
received the B.S. degree in Computer Science and Technology from the School of Computer Science, Nankai University, Tianjin, China, in 2025, where he is currently pursuing the M.S. degree. 

His research interests include computer vision and 3D reconstruction, with a particular focus on underwater vision.
\end{IEEEbiography}

\vspace{-1cm}

\begin{IEEEbiography}
[{\includegraphics[width=1in,height=1.25in,clip,keepaspectratio]{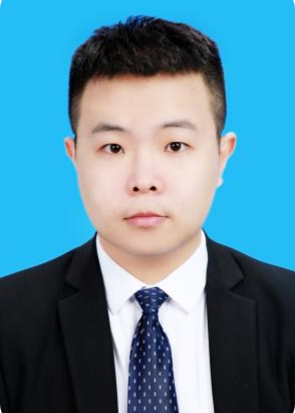}}]{Yuanlin Zhang}
received his B.S. degree from Tianjin University of Technology, Tianjin, China, in 2021, and his M.S. degree from Minzu University of China, Beijing, China, in 2025. From 2022 to 2024, he was a joint training student at the Institute of Software, Chinese Academy of Sciences, Beijing, China. He is currently a research intern at Nankai University, Tianjin, China. 

His research interests include image restoration and enhancement and multimodal learning.
\end{IEEEbiography}

\vspace{-5pt}
\begin{IEEEbiography}
[{\includegraphics[width=1in,height=1.25in,clip,keepaspectratio]{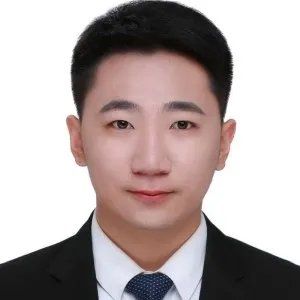}}]{Chunle Guo}
(Member, IEEE) received the Ph.D. degree from Tianjin University, China, under the supervision of Prof. Ji-Chang Guo. He was a Visiting Ph.D. Student with the School of Electronic Engineering and Computer Science, Queen Mary University of London (QMUL), U.K. He was a Research Associate with the Department of Computer Science, City University of Hong Kong (CityUHK). He was a Postdoctoral Researcher with Nankai University, under the guidance of Prof. Ming-Ming Cheng. He is currently an Associate Professor with Nankai University. 

His current research interests include image processing, computer vision, and deep learning, particularly in the domains of image restoration and enhancement.
\end{IEEEbiography}

\vspace{-5pt}
\begin{IEEEbiography}
[{\includegraphics[width=1in,height=1.25in,clip,keepaspectratio]{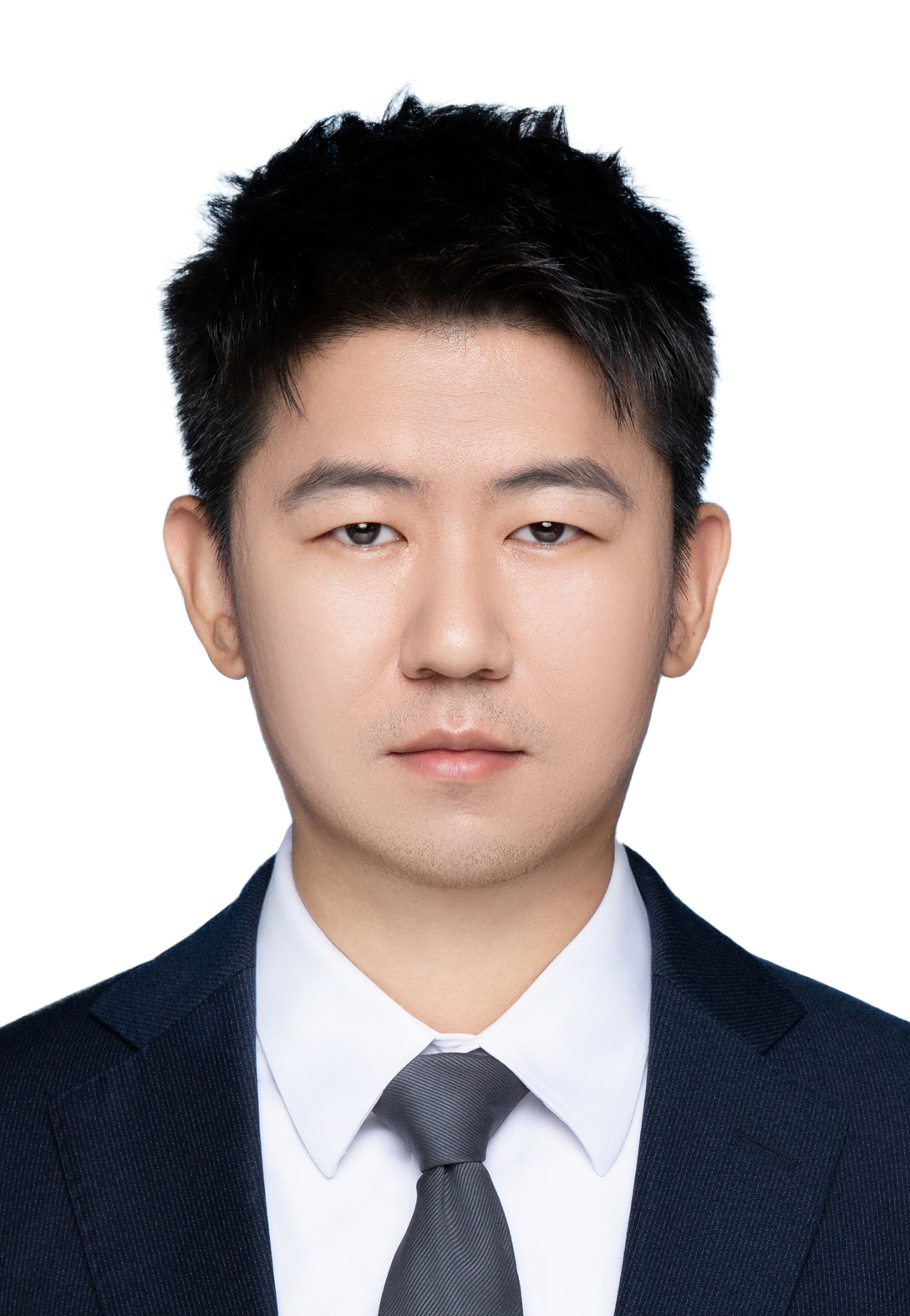}}]{Ruixing Wang}
is currently at the camera group of DJI. He received the B.S. degree from Huazhong University of Science and Technology in 2016, and the Ph. D. degree from the Chinese University of Hong Kong, in 2021. Before joining in DJI, he was a principal engineer at Honor Device Co., Ltd. He serves as a reviewer for CVPR, ICCV, ECCV, NeurIPS, ICML, ICLR, AAAI, WACV, ACCV, TPAMI, IJCV, etc. 

His research interests include computational photography and image processing. 
\end{IEEEbiography}
\vspace{-5pt}
\begin{IEEEbiography}
[{\includegraphics[width=1in,height=1.25in,clip,keepaspectratio]{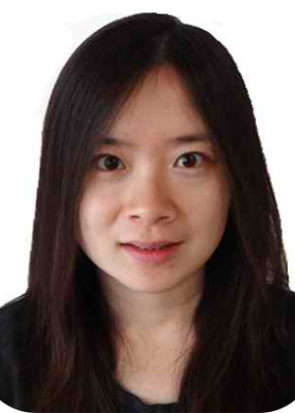}}]{Xiongxin Tang}
received her Ph.D. degree from the University of Chinese Academy of Sciences in 2013. She is currently an Associate Professor at the Institute of Software, Chinese Academy of Sciences. With extensive experience in optical simulation calculations and optical software development, she has actively engaged in research and development in this field. She has made significant contributions to her field with over 20 published academic papers in reputable domestic and international journals and conferences. 
\end{IEEEbiography}
\vspace{-5pt}
\begin{IEEEbiography}
[{\includegraphics[width=1in,height=1.25in,clip,keepaspectratio]{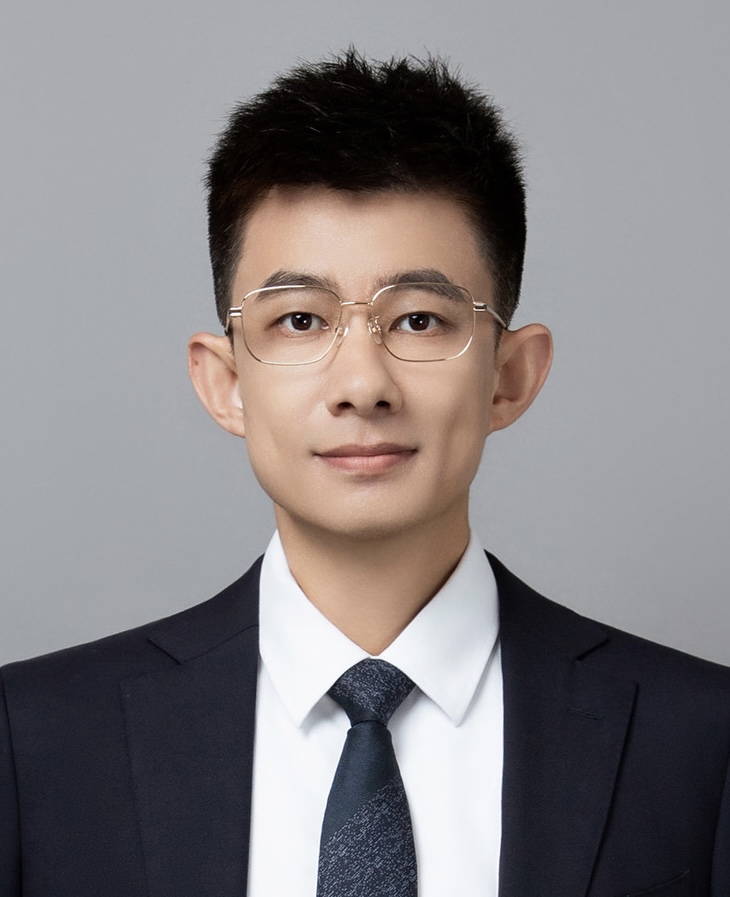}}]{Chongyi Li}
(Senior Member, IEEE) received the Ph.D. degree from the School of Electrical and Information Engineering, Tianjin University, Tianjin, China, in June 2018. From 2016 to 2017, he was a Joint Training Ph.D. Student with The Australian National University, Australia. He was a Research Fellow with the City University of Hong Kong and Nanyang Technological University from 2018 to 2021. He was a Research Assistant Professor with the School of Computer Science and Engineering, Nanyang Technological University, from 2021 to 2023. He is currently a Full Professor with the School of Computer Science, Nankai University, China. 

His current research interests
include image processing, computer vision, and deep learning, particularly in
the domains of image restoration and enhancement.
\end{IEEEbiography}

\vfill

\end{document}